\pdfoutput=1
\documentclass[11pt,breaklinks,hidelinks]{article}

\usepackage[final]{acl}
\usepackage{times}
\usepackage{latexsym}
\usepackage[T1]{fontenc}
\usepackage[utf8]{inputenc}
\usepackage{microtype}
\usepackage{inconsolata}
\usepackage{booktabs}
\usepackage[table,xcdraw]{xcolor}
\usepackage{tabularx}
\usepackage{makecell}
\usepackage{multirow}

\usepackage{lscape}  % 引入 landscape 环境
\usepackage{times}
\usepackage{xcolor}
\usepackage{soul}
\usepackage{url}
\usepackage{graphicx}
\usepackage{amsmath}
\usepackage{amsthm}
\usepackage{booktabs}
\usepackage{algorithm}
\usepackage{algorithmic}
\usepackage[switch]{lineno}
\usepackage{amsfonts}
\title{FinEval-KR: A Financial Domain Evaluation Framework for Large Language Models' Knowledge and Reasoning}

\author{
Shaoyu Dou\thanks{Equal contribution.}  \\ Ant Group \And
Yutian Shen$^{\ast}$\thanks{This work was completed during the internship at Ant Group.} , Mofan Chen$^{\ast \dagger}$  \\ {\bf Zixuan Wang$^{\dagger}$ \and Jiajie Xu$^{\dagger}$ } \\ Shanghai University of Finance and Economics
\AND
Qi Guo, Kailai Shao \\ {\bf  Chao Chen \and Haixiang Hu } \\ Ant Group \And
Haibo Shi \\ {\bf Min Min \and Liwen Zhang\thanks{Corresponding author.} } \\ Shanghai University of Finance and Economics  
}

\begin{document}
\maketitle
\begin{abstract}
Large Language Models (LLMs) demonstrate significant potential but face challenges in complex financial reasoning tasks requiring both domain knowledge and sophisticated reasoning. Current evaluation benchmarks often fall short by not decoupling these capabilities indicators from single task performance and lack root cause analysis for task failure. To address this, we introduce FinEval-KR, a novel evaluation framework for decoupling and quantifying LLMs' knowledge and reasoning abilities independently, proposing distinct knowledge score and reasoning score metrics. Inspired by cognitive science, we further propose a cognitive score based on Bloom's taxonomy to analyze capabilities in reasoning tasks across different cognitive levels. We also release a new open-source \textbf{Chinese} financial reasoning dataset covering 22 subfields to support reproducible research and further advancements in financial reasoning. 
Our experimental results reveal that LLM reasoning ability and higher-order cognitive ability are the core factors influencing reasoning accuracy. We also specifically find that even top models still face a bottleneck with knowledge application. Furthermore, our analysis shows that specialized financial LLMs generally lag behind the top general large models across multiple metrics.
\end{abstract}

\section{Introduction}

In recent years, rapid LLM development has led the transformation of artificial intelligence. LLMs demonstrate strong natural language processing capabilities and inspire application innovations across various fields, including scientific research~\cite{zhang2024comprehensive}, financial services~\cite{nie2024survey}, content creation~\cite{betker2023improving}, etc.
However, achieving satisfactory performance for complex tasks, such as financial decision making, proves difficult when relying solely on knowledge introduced during training or instructions~\cite{wang2025can,liu2025fin,wang2024re}. 
Reasoning ability, that is, the ability of logical deduction, problem solving, and abstract thought, stands as a core hallmark of advanced intelligence and becomes crucial for assessing LLM intelligibility and applicability~\cite{wang2024mars,li2024llms,valmeekam2023planbench}.
Therefore, it is critical to accurately assess the knowledge and reasoning abilities of LLMs. This helps to understand the shortcomings of the model to support targeted optimization.

% --------------
Although several LLM evaluation benchmarks have been proposed, they still have several limitations in evaluating reasoning capabilities.

\textit{Insufficient capability decoupling}. Mainstream benchmarks typically evaluate a model's capabilities based on its performance across various tasks. Their performance represents either the model's knowledge capacity~\cite{nie2024cfinbench,liu2024mtfineval} or its reasoning ability~\cite{DBLP:conf/iclr/Saparov023,geva2021did}. However, our experiments demonstrate that LLM performance in reasoning tasks is influenced by both knowledge and reasoning ability. Therefore, it is essential to decouple and quantify them separately to achieve a more accurate capability characterization.
    
\textit{Lack of root cause analysis}. Current reasoning evaluation frameworks mostly focus on the  correctness of the reasoning processes and results, while neglecting the diagnosis of the erroneous result. Specifically, there is not yet an effective way to distinguish whether a reasoning failure stems from knowledge gaps, such as unclear concept comprehension, or from flaws in the reasoning processes, like missing reasoning steps. This limits the potential for specifically optimizing the model.

\textit{Neglect of the cognitive science perspective in financial LLM benchmark}. While some benchmarks have begun to assess reasoning ability, they generally lack a design grounded in cognitive science--a critical omission for the financial domain. Unlike the deductive reasoning of law or the diagnostic processes of medicine, the essence of financial decision-making is a \textit{quantitative game against future uncertainty}. This requires a spectrum of higher-order cognitive abilities that extend far beyond simple knowledge application~\cite{zhang2025xfinbench}. For instance, evaluating the impact of a central bank's interest rate policy requires not only \textit{applying} knowledge of rate-to-exchange dynamics but also \textit{evaluating} the complex interplay between market sentiment and strategic policy expectations. This distinction underscores the need for a framework like Bloom's Taxonomy as an essential diagnostic tool to pinpoint the specific cognitive deficiencies that hinder advanced financial reasoning in LLMs.

The above limitations motivate us to explore the following questions:

\textbf{Q1:} How do knowledge and reasoning ability jointly determine the performance of LLMs in domain reasoning tasks?

\textbf{Q2:} Can we develop an evaluation framework that decouples and independently quantifies knowledge and reasoning ability from task performance?

We begin by empirically verifying the fundamental role of knowledge in domain-specific reasoning through a preliminary experiment in the finance sector. Building on these initial findings, we make the following contributions:
\begin{itemize}
    \item \textbf{A novel evaluation framework}. We propose a novel LLM evaluation framework that disentangles the assessment of domain knowledge and reasoning ability from task performance metrics. This allows us to introduce two distinct metrics: a \textit{Knowledge Score} and a \textit{Reasoning Score}. Furthermore, inspired by cognitive science, we posit that complex reasoning relies on a hierarchy of cognitive abilities. This motivates our third metric, the \textit{Cognitive Score}, which leverages Bloom's Taxonomy to provide a fine-grained analysis of the cognitive processes employed by LLMs during reasoning.

    \item \textbf{A new open-source Chinese-language dataset for financial reasoning}\footnote{\url{https://github.com/SUFE-AIFLM-Lab/FinEval-KR}}. The released dataset encompasses 22 key financial subfields, its primary contribution lies in its multi-layered annotations, where each sample includes knowledge point labels, step-by-step reasoning chains, and the required cognitive skills. As a comprehensive and deeply annotated resource, it is designed to serve as a specialized benchmark to advance research in both financial and cognitive reasoning.

    \item \textbf{Evaluation results of mainstream LLMs}. Based on the proposed framework and dataset, we conduct a comprehensive evaluation of the current mainstream LLMs, which verifies the effectiveness of the proposed evaluation methodology and yields a series of insightful conclusions (see Section~\ref{sec:results} and Appendix~\ref{sec:app_experiment} for details).
\end{itemize}

\section{Related Work}

Recent studies highlight a paradigm shift in LLM evaluation, moving from task-specific benchmarks to capability-based assessments of core competencies like knowledge and reasoning~\cite{cao2025toward}. Financial LLM benchmarks have followed a similar trajectory, evolving through three main phases. Early benchmarks adapted general NLP tasks to the financial domain (e.g., FinGPT~\cite{wang2023fingpt}, CFBenchmark~\cite{lei2023cfbenchmark}). As tasks grew more complex, the focus shifted to specialized knowledge evaluation (e.g., FinEval~\cite{zhang2023fineval}, FinTruthQA~\cite{xu2024fintruthqa}). Contemporary benchmarks now incorporate complex decision-making tasks requiring integrative reasoning, such as market analysis and risk assessment (e.g., InvestorBench~\cite{li2024investorbench}, FinBen~\cite{xie2024finben}).

Despite this progress, financial reasoning poses unique challenges that current evaluation methods struggle to address. It demands (1) comprehension of complex, multi-modal data (e.g., time-series and unstructured text), (2) deep domain-specific knowledge, and (3) advanced computational and deductive skills. This complexity creates a critical limitation in existing benchmarks: overall performance metrics entangle knowledge mastery with reasoning ability, making it impossible to diagnose the true source of a model's failure. This fundamental challenge motivates our primary contribution: a decoupled assessment framework. Furthermore, these unique demands directly guided the design of our financial reasoning dataset (see Section \ref{sec:dataset}).

Beyond the issue of entangled evaluation, a second key limitation exists: the lack of fine-grained cognitive analysis. While cognitive science perspectives offer crucial insights for LLM optimization~\cite{DBLP:conf/coling/HuberN25,adams2015bloom}, they are largely overlooked in financial benchmarks. This gap hinders targeted model improvement, as engineers cannot pinpoint specific cognitive deficiencies (e.g., analysis vs. evaluation) to address during fine-tuning. Our work fills this gap by introducing a cognitive evaluation dimension, enabling a more diagnostic approach to model development.

\section{Preliminary Experiment}
To further illustrate our research motivation, we design a set of preliminary experiments focusing on a simple reasoning task in the financial domain -- a financial calculation problem that requires only a single formula. 
For the dataset and prompts used in this experiments, please refer to Appendix~\ref{sec:app_pri_prompts}.

\subsection{Experiment Settings}
In order to correctly solve such problems, LLMs need to complete the following three sub-tasks:
(1) Recall the calculation formula for the variable to be solved.
(2) Identify the variable names and their values from the problem statement.
(3) Substitute the variable values into the formula and calculate the target variable.
According to Bloom's Taxonomy, these three sub-tasks correspond to remembering, understanding and applying/analyzing respectively. Obviously, the model can only answer the questions correctly if all these subtasks are done correctly.

We design the following three comparative experiments to evaluate the impact of knowledge on reasoning task:
\textit{Experiment 1} (E1). The LLM is directly prompted with the original question and asked to complete the entire reasoning process independently.
\textit{Experiment 2} (E2). Based on experiment 1, inject variables and their values into the prompt.
\textit{Experiment 3} (E3). Further provide formulas for calculation based on experiment 2.

To ensure the fairness, we add an equal amount of irrelevant information as distractors to the control groups, while providing the key knowledge in the experimental groups. The experiments are first conducted on the Qwen2.5-7B\_Instruct, and the generalizability of the findings is verified using GPT-4o. The experimental results are presented in Table~\ref{tab:preliminary}.

\begin{table}[!bth]
\centering
\scalebox{0.8}{
\begin{tabular}{@{}cccc@{}}
\toprule
Model/Settings & E1     & E2     & E3     \\ \midrule
Qwen2.5-7B\_Instruct     & 58.0\% & 72.4\% & 85.9\% \\
GPT-4o         & 64.5\% & 80.5\% & 92.5\% \\ \bottomrule
\end{tabular}
}
\caption{Cumulative correctness rate of reasoning task in three experimental settings.}
\label{tab:preliminary}
\end{table}

\subsection{Results Analysis}
Experiment 3 to 1 can be regarded as knowledge stripping experiments. As key knowledge is progressively removed, the problem-solving rate decreases significantly, suggesting that the lack of knowledge is often the root cause of reasoning failure in reasoning tasks. This leads to the following conclusion.

\textit{In complex domain reasoning tasks, knowledge is necessary for successful reasoning.}

On the contrary, Experiments 1 to 3 can be regarded as knowledge enhancement experiments, and the results show that the reasoning success rate of LLM is significantly improved after the introduction of key knowledge in the prompts. However, even knowledge is sufficiently injected into the prompt, GPT-4o still persists with an error rate of about 7.5\%, suggesting that reasoning ability may become a performance bottleneck for such tasks. Combined with the previous conclusion, we infer that:

\textit{Knowledge is a necessary but not sufficient condition for successful reasoning.}

It is noteworthy that both models exhibit significant knowledge dependence in all experimental settings, while the performance gap between them persists. 
This performance discrepancy suggests that there may be significant differences in the knowledge and reasoning capabilities of different models. Therefore, the decoupled evaluation framework can help us identify more clearly the shortcomings of the models in terms of knowledge and reasoning capabilities.

From a cognitive science perspective, knowledge stripping experiments revealed the damaging effects of lower-order cognitive deficits on higher-order reasoning, as evidenced by a 27.94\% and 28\% decrease in reasoning accuracy in qwen2.5-7b and GPT-4o, respectively. This change also supports the progressive dependence between cognitive levels. 
Furthermore, the performance difference between the two models in the same experiment settings suggests that there is a significant gap between them, at least in the remembering layer.

\section{FinEval-KR}

In this section, we present a \textbf{Fin}ancial domain \textbf{Eval}uation framework for assessing \textbf{K}nowledge and \textbf{R}easoning abilities (\textbf{FinEval-KR}). We first describe the methodology used to construct the evaluation dataset. Next, we introduce a multi-stage evaluation framework that performs root cause analysis of reasoning errors via knowledge-augmented question answering, enabling a decoupled assessment of a model’s knowledge mastery and reasoning ability.
Finally, we define a series of evaluation metrics: a \textit{knowledge score} based on domain knowledge coverage; a \textit{reasoning score} and a \textit{cognitive score} which is based on Bloom's taxonomy. This approach improves the interpretability of LLM evaluations in financial scenarios and provides clear directions for targeted model improvement.

\subsection{Benchmark Dataset Construction}
\label{sec:dataset}
The FinEval-KR benchmark dataset is constructed through four steps: data collection and processing, automated question generation, answer generation, and dataset annotation. This framework ensures comprehensive coverage of financial knowledge domains while maintaining academic rigor. All prompt templates for the dataset generation are detailed in Appendix~\ref{sec:app_dataset_generation}.

\paragraph{Corpus Collection}
To ensure the dataset is both authoritative and relevant, we selected nine canonical textbooks from major financial disciplines. These sources provide a comprehensive and up-to-date overview of modern finance. This process yielded a financial corpus totaling 8,460 pages. A detailed list of the textbooks and the rationale for their selection is available in Appendix~\ref{sec:app_datasource}.

\paragraph{Question Generation}
We generate financial problems from the obtained corpus using a two-stage process. First, we use a custom-designed prompt to instruct OpenAI o1, a state-of-the-art model at the time of our research that was specially enhanced for reasoning capabilities, to create a computational question based on a given text segment. The prompt's design ensures the question meets predefined standards (see Figure~\ref{fig:generate_question} in the appendix for details). Second, we subject each candidate question to a three-step automated validation: it is checked for logical coherence, consistency with the source material, and overall quality. Questions that fail any check are discarded, ensuring the high fidelity of the final dataset. The details of validation please refer to Appendix~\ref{sec:app_validation}.

\paragraph{Answer Generation}
We generate and validate the ground-truth answers using a structured three-stage pipeline:
(1) Unconstrained generation: We first prompt OpenAI o1 to solve each problem, guided by the prompt shown in Figure~\ref{fig:generate_answer}. Crucially, we impose no initial format constraints on the output, a strategy designed to capture the model's most natural and diverse problem-solving pathways.
(2) Standardized formatting: The resulting raw solutions are then systematically parsed and reformatted according to a predefined template to ensure consistent and uniform presentation across the dataset.
(3) Rigorous validation: Finally, each formatted answer undergoes a triple-validation protocol, which mirrors the question validation process (see Appendix~\ref{sec:app_validation} for details).

\paragraph{Dataset Annotation}
For each question, we use OpenAI o1 guided by the prompt in Figure~\ref{fig:annotate_knowledge_point} to identify all requisite knowledge points. These include, but are not limited to, core financial concepts, regulatory frameworks and mathematical operations. 

For each step in the answer, we annotate the corresponding cognitive level using Bloom's Taxonomy. To mitigate the inherent subjectivity of this task, we developed a constrained, keyword-driven methodology. We guide the OpenAI o1 using a predefined set of keywords strongly associated with each cognitive level, derived from ~\citet{anderson2001taxonomy} (see Figure~\ref{fig:annotate_cognitive_label} for prompt template). This process maps each reasoning step to one or more levels: Remembering, Understanding, Applying, Analyzing, and Evaluating.
We intentionally exclude the Creating level. This decision ensures objectivity, as our dataset comprises problems with determinate solutions, a principle that aligns with standardized financial certification exams.

\begin{figure*}[!bt]
	\centering
	\includegraphics[width=0.98\linewidth]{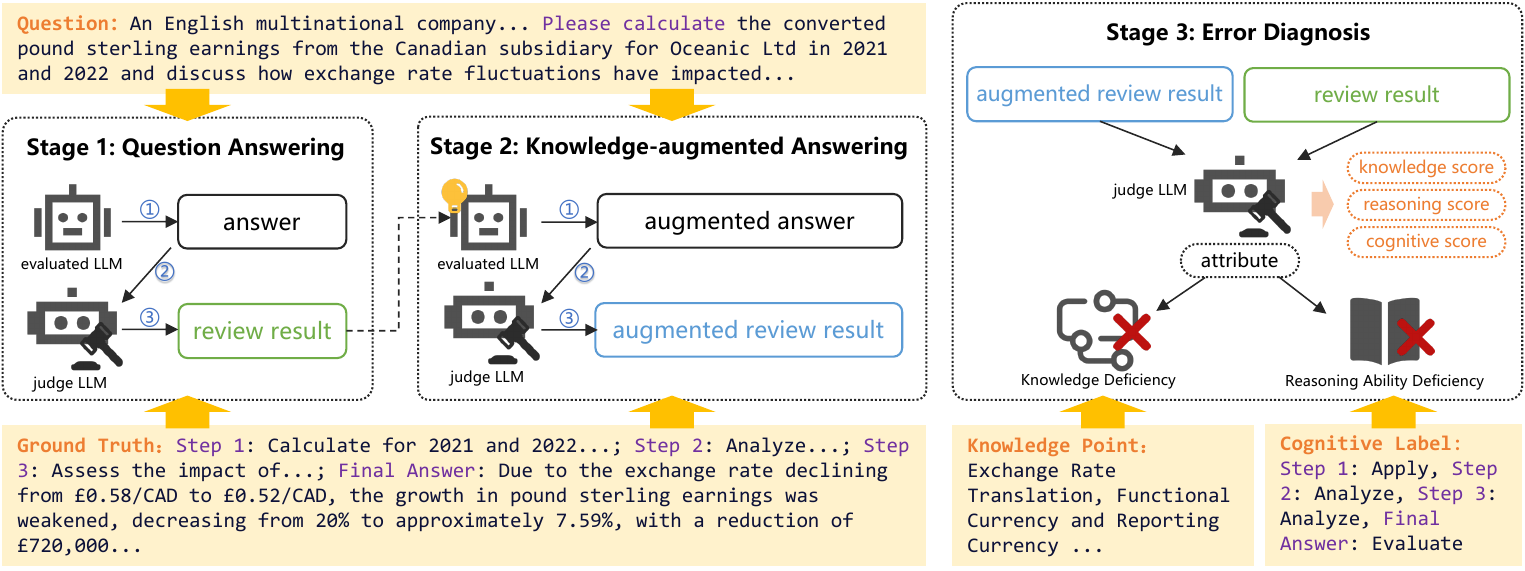}
	\caption{Three-stage evaluation framework of FinEval-KR, and an exemplary sample of the dataset. Note that the original dataset is in Chinese, the figure provides an English translation for readability.}
	\label{fig:framework}
\end{figure*}

\paragraph{Dataset Characteristics and Exemplary Samples}
The final evaluation set contains a total of 9,782 question-answer pairs and their associated annotations. The questions and answers in the dataset are verified by human experts, and a sample check showed that the accuracy of the dataset is above 90\%. Further details on the human annotators, the quality control mechanisms, and the validation process are provided in the Appendix~\ref{sec:app_labelor}.
An exemplary sample from the constructed dataset is given in Figure~\ref{fig:framework} on a yellow background. (see Figure~\ref{fig:full_example} for the complete example).
The complete statistical characterization of the dataset is detailed in Appendix~\ref{sec:appendix_stat}.

% -------

\subsection{Evaluation Framework}

As shown in Figure~\ref{fig:framework}, the FinEval-KR framework comprises three evaluation stages.
First, a model attempts a problem, and an LLM-as-a-judge identifies any errors, extracting the specific knowledge points needed for a correct solution. In the second stage, we provide the model with this missing knowledge and have it re-answer the question. The final stage performs a comparative analysis for attributing the initial error to knowledge deficiency or reasoning ability deficiency. This mechanism allows for the independent quantification of a model's knowledge and reasoning capabilities. 

\subsubsection{Stage 1: Question Answering}
\paragraph{Unconstrained Solution Generation}
In this initial stage, the model under evaluation generates a solution without any format constraints. This design is crucial for preventing judgment errors based on superficial format-matching. It ensures, for example, that a logically sound reasoning path is not unfairly penalized simply for deviating from the step-order of the reference answer. The prompt for this stage is detailed in Appendix~\ref{sec:app_framework}, Figure~\ref{fig:question_answering}.

\paragraph{Structured Judgment and Error Analysis}
Next, we employ an LLM-as-a-judge to analyze the generated solution against a reference answer. To ensure the judgment is objective and reproducible, the judge is guided by a highly structured, Chain-of-Thought (CoT) prompt that enforces a rigorous step-by-step analysis (see Figure~\ref{fig:judge}). Additionally, we address the potential bias challenge of the judge model in Appendix~\ref{sec:judge_bias}.

\begin{figure}[!bthp]
	\centering
	\includegraphics[width=0.98\linewidth]{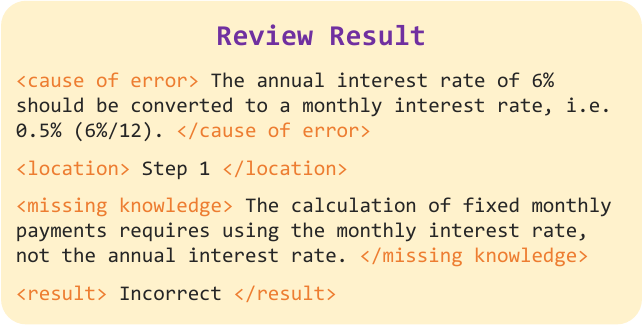}
	\caption{Example of review result generated by the judge model (original in Chinese, with English translation).}
	\label{fig:review_result}
\end{figure}

If an error is detected, the judge's output, termed the \textit{review result}, pinpoints the first incorrect step, identifies its root cause, and lists the corresponding knowledge deficiencies (an example is shown in Figure~\ref{fig:review_result}).
If the final answer is correct, we consider the entire reasoning process valid. This assumption is grounded in the novelty and multi-step complexity of our dataset, which makes correct answers via guessing or exploiting artifacts (the ``Clever Hans'' effect) highly improbable. 

\subsubsection{Stage 2: Knowledge-augmented Answering}
If the evaluated model makes a error in Stage 1, the framework proceeds to this second stage. Here, we provide the model with the exact knowledge points that the judge identified as missing in Stage 1. This knowledge is integrated into a new prompt (see Appendix~\ref{sec:app_framework}, Figure~\ref{fig:re-answer}), instructing the model to re-attempt the problem.
The core purpose of this stage is to isolate the reasoning variable. By explicitly providing the necessary knowledge -- a prerequisite for correct reasoning as validated in our preliminary study -- we can now assess if the model can reason correctly when its knowledge gaps are filled. The judge then re-evaluates the new solution using the same protocol as in Stage 1. The outcome of this assessment is termed the \textit{augmented review result}.

\subsubsection{Stage 3: Error Diagnosis}
The final stage performs a comparative analysis between the outcomes of Stage 1 and Stage 2 to determine the root cause of the initial error. 
Our approach is grounded in the principle that \textit{LLM's preference to external information reveals its internal knowledge gaps}~\citep{Wu2024How}. Specifically, the judge model compares the \textit{review result} with the \textit{augmented review result}, following the principles below to determine the root cause of reasoning errors:
\begin{itemize}
    \item \textit{Knowledge Deficiency}. If the augmented review result shows that the model reasons correctly in Stage 2, or the erroneous step occurs later than in Stage 1. This indicates that the evaluated model preferred the augmented knowledge in the second stage. This proves that the initial error is caused by knowledge deficiency.
    
    \item \textit{Reasoning Ability Deficiency}. If the evaluated model still makes a reasoning error in Stage 2, and the erroneous step is consistent with Stage 1, this indicates that the evaluated model still preferred its internal prior knowledge. This proves that the initial error is rooted in poor reasoning ability.
\end{itemize}

Through this attribution method, our framework successfully decouples a model's knowledge and reasoning abilities from its overall task performance. 

\subsubsection{Evaluation Metrics}
We propose three core metrics: knowledge score, reasoning score, and cognitive score, while also retaining accuracy to measure the model's overall task performance.

\paragraph{Knowledge Score (KS)} This metric quantifies the evaluated model's knowledge coverage in the financial domain.
\begin{equation}
    \label{equ:ks}
    KS  = 1 - \frac{\left| \bigcup_{i=1}^{M} K'_i \right|}{\left| \bigcup_{i=1}^{N} K_i \right|},
\end{equation}
where $M$ is the number of errors attributed to knowledge deficiency during evaluation. $K'_i$ denotes the set of knowledge points involved in the erroneous reasoning steps for the $i$-th question whose error was attributed to a knowledge deficiency. $N$ is the total number of evaluation samples. $K_i$ denotes the set of knowledge points involved in the $i$-th evaluation question. The denominator in the Eq.~(\ref{equ:ks}) is the total number of knowledge points across all evaluation questions in the dataset.

\paragraph{Reasoning Score (RS)} This metric measures the evaluated model's reasoning ability in the financial domain.
\begin{equation}
    \label{eqn:RS}
    RS = \frac{\sum_{i=1}^{N} \mathbb{I}(a_i=a_{i}^{\text{ref}})}{N- \sum_{i=1}^{N} \mathbb{I}(a_i \neq a_{i}^{\text{ref}} \land r(a_i) = \mathrm{K}) },
\end{equation}
where $a_i$ is the answer of the evaluated model for the $i$-th question, and $a_{i}^{\text{ref}}$ is the corresponding reference answer. Thus, the numerator of Eq.~(\ref{eqn:RS}) is the total number of questions with correct reasoning. $r(a_i)$ represents the root cause for the erroneous result $a_i$, which can be either $\mathrm{K}$ or $\mathrm{R}$, representing knowledge and reasoning ability deficiency, respectively. 
In addition, since each reasoning step in the dataset is annotated with a cognitive label, the reasoning ability deficit can be further subdivided into the ability deficit at a certain cognitive level, i.e., $\mathrm{R}^{j}$, $j \in \{1,2,3,4,5\}$, where $j$ is the index of the level in Bloom's taxonomy, and $1,2,3,4,5$ denote remembering, understanding, applying, analyzing, and evaluating respectively. The denominator in Eq.~(\ref{eqn:RS}) is the total number of questions whose reasoning errors are not attributed to knowledge deficiencies.

\paragraph{Cognitive Score (CS)} Furthermore, to explore the level of cognition exhibited by LLMs when solving reasoning tasks, we expanded a series of fine-grained metrics drawing on RS. The $j$-th cognitive level score of the LLM is defined as,
\begin{equation}
\begin{aligned}
\label{eqn:rs_j}
    & CS_j = RS \times (1- \\
    & \alpha_j \times \frac{\sum_{i=1}^{N} \mathbb{I}    (r(a_i)=\mathrm{R}^{j})}{N- \sum_{i=1}^{N} \mathbb{I}(a_i \neq a_{i}^{\text{ref}} \land r(a_i) = \mathrm{K})} ).
\end{aligned}
\end{equation}
$\alpha_j \in (0,1)$ is a penalty coefficient, and it is designed to have negative correlation with the cognitive level $j$. This weighting scheme is designed to heavily penalize errors made by the model during lower-level cognitive reasoning steps. In our experiments, we empirically set $\alpha_1, \cdots, \alpha_5$ to a linearly decreasing sequence 0.9, 0.8, 0.7, 0.6 and 0.5.

\paragraph{Accuracy (Acc)} Finally, the success rate for all reasoning tasks is defined as,
\begin{equation}
    Acc = \frac{1}{N}\sum_{i=1}^{N} \mathbb{I}(a_i=a_{i}^{\text{ref}}).
\end{equation}

\section{Experiments}
This section first validates the FinEval-KR framework's effectiveness in identifying potential knowledge weaknesses and performing root cause analysis. Subsequently, we selected a range of LLMs, and evaluated them using the proposed FinEval-KR framework and our open-sourced dataset.

\subsection{Alignment Analysis of Judge Model in FinEval-KR}
\label{sec:effectiveness}
The core of FinEval-KR lies in the design of the judge model for identifying initial reasoning step, recalling potential knowledge weaknesses and pinpointing root causes, which is fundamental for achieving decoupled evaluation. 
To this end, we design a set of comparative experiments to assess the alignment between the judge model and human evaluation. 

The methods participating in the comparison include:
(a) \textit{Direct Prompting}: We use advanced models like OpenAI o1, directly prompting them to execute knowledge identification and root cause localization tasks.
(b) \textit{Task Decomposition}: Decompose the above two tasks into multiple subtasks and clearly define the logical dependencies between these subtasks
(c) \textit{Ours}: Based on task decomposition, this method requires the model to self-reflect and explicitly output reasoning processes step-by-step before generating the final conclusion (i.e., adding the \texttt{<Inner Thought>} as shown in Figure~\ref{fig:judge}).

Given that our benchmark dataset is composed in Chinese, we select Qwen2.5-72B\_Instruct, which is specifically optimized through extensive pre-training on Chinese corpora, as backbone LLM in the latter two methods. We use the following three metrics to evaluate the performance of these methods:
(a) \textit{Accuracy of error identification}: This refers to the proportion of correctly localized initial reasoning step in the Stage 1.
(b) \textit{Accuracy of recalled knowledge points}: The percentage of missing knowledge correctly recalled in Stage 1.
(c) \textit{Accuracy of error attribution }: The proportion of correctly attributed reasoning errors to knowledge or reasoning ability deficits.

All metrics are calculated after a manual review of the model outputs by financial domain experts. Experimental results are presented in Table~\ref{tab:framework_effectiveness}. It can be observed that FinEval-KR outperforms the comparison methods across all metrics, demonstrating its superior performance in error attribution and knowledge identification tasks. 

Additionally, we discuss the limitations of adopting Qwen2.5-72B\_Instruct as the judge model in the Limitations section. 

\begin{table}[!tbp]
\centering
\scalebox{0.8}{
\begin{tabularx}{1.25\linewidth}{cccc}
    \toprule
    Methods & \makecell[c]{Error\\ Identification} & \makecell[c]{Knowledge\\Recall} & \makecell[c]{Error\\Attribution} \\ \midrule
    \makecell[c]{Direct\\Prompting} &  0.56&  0.24&  0.20 \\
    \makecell[c]{Task\\Decomposition} &  0.85&  0.60&  0.53 \\
    Ours &  \textbf{0.92}&  \textbf{0.94}&  \textbf{0.93} \\
    \bottomrule
\end{tabularx}
}
\caption{Accuracy for three evaluation tasks.}
\label{tab:framework_effectiveness}
\end{table}

\begin{table*}[t]
\centering
\scalebox{0.85}{
\begin{tabular}{@{}ccccccccc@{}}
\toprule
Model/Metrics          & Acc             & KS              & RS              & \makecell[c]{CS$_1$\\(remember)} & \makecell[c]{CS$_2$\\(understand)} & \makecell[c]{CS$_3$\\(apply)}   & \makecell[c]{CS$_4$\\(analyze)} & \makecell[c]{CS$_5$\\(evaluate)} \\ \midrule
Qwen2.5-14B\_Instruct  & 0.5473          & 0.8490          & 0.6863          & 0.6547           & 0.6603             & 0.3893          & 0.6863          & 0.6820           \\ 
QwQ-32B-preview        & 0.7380          & 0.9073          & 0.8627          & 0.8450           & 0.8503             & 0.6987          & 0.8510          & 0.8597           \\ 
DeepSeek-v3            & 0.8270          & 0.9427          & 0.9077          & 0.8963           & 0.8993             & 0.7963          & 0.8943          & 0.9057           \\ 
DeepSeek-R1          & 0.8700          & 0.9517          & \textbf{0.9347} & \textbf{0.9377}  & \textbf{0.9397}    & \textbf{0.8810} & \textbf{0.9380} & \textbf{0.9433}  \\ 
Doubao-pro-32k         & 0.7825          & 0.9195          & 0.8750          & 0.8560          & 0.8600              & 0.7340          & 0.8565          & 0.8720          \\
Moonshot-v1-128k       & 0.4533          & 0.8340          & 0.6020          & 0.5620           & 0.5670             & 0.2763          & 0.5653          & 0.5973           \\
Ernie-bot-4.0          & 0.5733          & 0.8627          & 0.7053          & 0.6680           & 0.6753             & 0.4383          & 0.6847          & 0.6927           \\
Qwen-max-latest        & 0.6467          & 0.8797          & 0.7733          & 0.7507           & 0.7547             & 0.5340          & 0.7440          & 0.7703           \\
GPT-3.5-turbo          & 0.2830          & 0.7527          & 0.3973          & 0.3527           & 0.3603             & 0.0900          & 0.3893          & 0.3970           \\
GPT-4o                 & 0.6853          & 0.9020          & 0.8067          & 0.7847           & 0.7890             & 0.5930          & 0.7870          & 0.8030           \\
GPT-4.1                & 0.8263          & 0.9520          & 0.9063          & 0.8957           & 0.8977             & 0.7890          & 0.8927          & 0.9050           \\ 
o1-mini                & 0.7503          & 0.8997          & 0.8453          & 0.8340           & 0.8363             & 0.6983          & 0.8477          & 0.8450           \\
o3-mini                & 0.8207          & 0.9260          & 0.9070          & 0.9047           & 0.9073             & 0.8127          & 0.9023          & 0.9120           \\
Gemini-2.5-pro         & \textbf{0.8750} & \textbf{0.9627} & 0.9233          & 0.9123           & 0.9163             & 0.8403          & 0.9050          & 0.9120           \\
Gemini-2.5-flash       & 0.8440          & 0.9540          & 0.9203          & 0.9103           & 0.9133             & 0.8307          & 0.9100          & 0.9177           \\
Claude-3.7-sonnet      & 0.7923          & 0.9390          & 0.8823          & 0.8663           & 0.8703             & 0.7433          & 0.8653          & 0.8803           \\ 
Xuanyuan-FinX1-preview & 0.5890          & 0.8687          & 0.7323          & 0.7063           & 0.7130             & 0.4610          & 0.7323          & 0.7300           \\
Fin-R1-7B              & 0.4153          & 0.7510          & 0.5570          & 0.5190           & 0.5277             & 0.2170          & 0.5570          & 0.5527           \\ \bottomrule
\end{tabular}
}
\caption{Reasoning Accuracy (Acc), Knowledge Score (KS), Reasoning Score (RS), and Cognitive Score (CS) of evaluated LLMs on the FinEval-KR. The complete results please refer to Table~\ref{tab:all_results} in the appendix.}
\label{tab:results}
\end{table*}

\subsection{Evaluation}
For our evaluation, we select 18 leading and representative LLMs, referencing prominent leaderboards like the Chatbot Arena\footnote{\url{https://openlm.ai/chatbot-arena/}}. This selection spans a diverse range of models, including open-source and closed-source systems, models of varying parameter scales, and different architectures such as dense and Mixture-of-Experts (MoE) (see Appendix~\ref{sec:app_models} for details).

In our analysis, we focus on the relative performance rankings (i.e., performance tiers) of these models rather than their absolute scores. This approach is designed to ensure the robustness of our findings. While an LLM-as-a-judge may have inherent systematic biases, such biases have a smaller impact on the relative ordering of models than on their absolute scores. 

\subsection{Results and Core Findings}
\label{sec:results}
Table~\ref{tab:results} lists the average metrics for all 18 models from three independent runs, conducted with a model temperature of 1. The complete results, including standard deviations, are presented in Table~\ref{tab:all_results}.

The analysis of the model performance echelons and discussion of the results are detailed in in Appendix~\ref{sec:app_experiment}. In summary, the comprehensive analysis of all evaluation metrics identifies the current Tier 1 models as DeepSeek-R1, Gemini-2.5-pro and Gemini-2.5-flash. These models typically have parameters in excess of a hundred billion and use the MoE architecture to optimize computational resources. Furthermore, they specifically optimize reasoning capabilities through methods like reinforcement learning, and demonstrate outstanding performance in knowledge coverage and the completeness of reasoning paths.

\paragraph{Bottleneck in Knowledge Applying Abilities}
Our analysis reveals that it is reasoning and specific cognitive skills, not merely knowledge, that truly drive performance in advanced LLMs. While the Knowledge Score (KS) and Reasoning Score (RS) are positively correlated, KS scores converge among top-tier models, indicating that sheer knowledge is no longer the primary performance bottleneck. Instead, RS shows a strong correlation with accuracy, establishing reasoning ability as crucial for success. A deeper cognitive analysis pinpoints the ability to apply knowledge (CS$_3$) as the critical differentiator, evidenced by a sharp drop in this metric, which directly degrades their reasoning and accuracy.

Crucially, this weakness is not confined to lower-tier models. Even top-tier LLMs exhibit a significant deficit in applying knowledge (CS$_3$) compared to their abilities in analyzing (CS$_4$) or evaluating (CS$_5$). For instance, GPT-4.1 scores 0.7890 in CS$_3$ versus 0.8927 in CS$_4$. This universal shortcoming underscores a fundamental limitation of current models: a profound difficulty in transferring theoretical knowledge to practical, real-world application.

\paragraph{The Dilemma of Financial LLMs}
Our evaluation establishes Xuanyuan-FinX1-preview as the leading specialized financial LLM, consistently outperforming counterparts like Fin-R1-7B. However, a more critical finding emerges when comparing it to state-of-the-art general LLMs. Despite its domain leadership, Xuanyuan-FinX1-preview exhibits a significant performance gap of above 20\%, a deficit that spans across financial complex reasoning, and higher-order cognitive skills.
We attribute this gap to the superior generalization capabilities of leading general LLMs, which are developed through pre-training on vast, multi-domain datasets and benefit from more rapid iteration cycles. This advantage allows them to achieve strong performance even in financial fields, highlighting the limitations of current financial LLMs in terms of data diversity and development velocity.

Consequently, we predict a dual-track future for developing high-performance financial LLMs. The first track involves  building upon state-of-the-art general foundation models to leverage their vast world knowledge and robust generalization. The second requires employing advanced fine-tuning techniques to distill and align these models for specialized financial reasoning, moving beyond a simple reliance on domain data.

\section{Conclusion}
This paper introduces FinEval-KR, a novel evaluation framework designed to decouple and assess the knowledge and reasoning abilities of LLMs in the financial domain, supplemented by a cognitive perspective and a new dataset. Our evaluation results indicate that reasoning and higher-order cognitive abilities are crucial for reasoning accuracy. Even top models encounter a bottleneck in knowledge application, and specialized financial models generally lag behind top general LLMs.

\section*{Limitations}
A potential limitation of this study lies in the choice of the judge model. Our primary experiments were conducted using Qwen-2.5-72B\_Instruct, which represented the state-of-the-art among publicly available models with strong Chinese support during our experimental phase in late 2024. With the rapid evolution of large language models, even more capable reasoning models like DeepSeek-R1 have since emerged.

To investigate the impact of this evolution, we performed a small-scale evaluation using DeepSeek-R1 as the judge model. The results revealed a clear performance-efficiency trade-off: while DeepSeek-R1 yielded a marginal accuracy improvement of approximately 3\%, it nearly doubled the inference time, posing significant challenges for large-scale evaluation. Crucially, we found that although the absolute scores of the evaluated models slightly increased, their relative rankings and the performance gaps between them remained highly consistent. Since our research focuses on the comparative performance of different methods, this consistency confirms the robustness of our conclusions.

Our future work will focus on enhancing the framework's robustness by incorporating reasoning LLMs and more diverse evaluation paradigms. Specifically, we plan to employ multiple, heterogeneous models for dataset generation and implement a cross-validated, multi-judge evaluation pipeline to minimize potential biases, raise the evaluation ceiling, and bolster the benchmark's overall reliability.

\section*{Acknowledgement}
We thank the anonymous reviewers for their helpful suggestions. This work was supported by Ant Group Research Fund.

\bibliography{myref}

\appendix

\section{The Dataset and Prompts Used in Preliminary Experiments}
\label{sec:app_pri_prompts}
\begin{figure}[!hbt]
    \centering
    \includegraphics[width=0.9\linewidth]{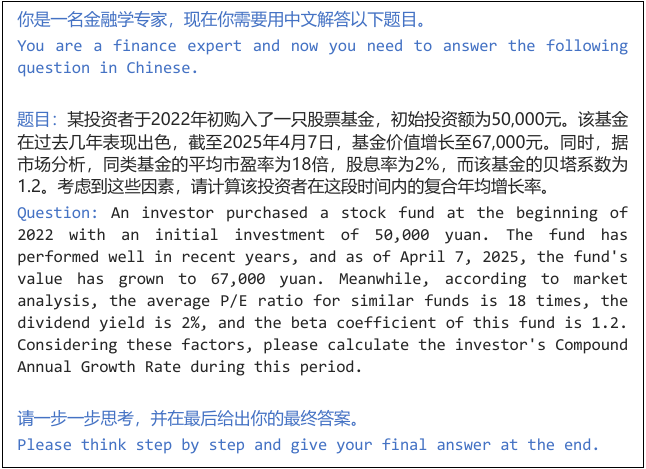}
    \caption{The prompt for experiment 1 and an exemplary sample (original in Chinese, with English translation).}
    \label{fig:app_pri_exp1}
\end{figure}

\begin{figure}[!hbt]
    \centering
    \includegraphics[width=0.9\linewidth]{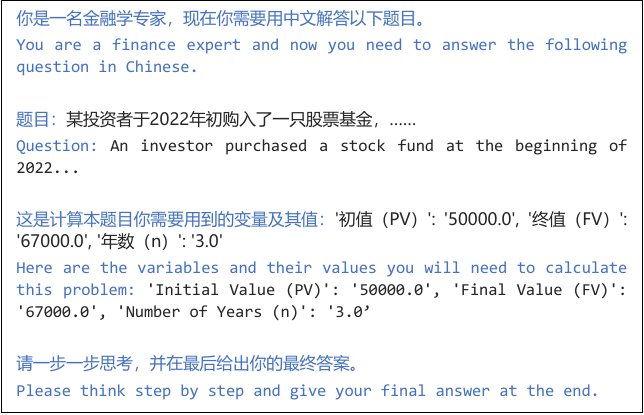}
    \caption{The prompt for experiment 2 and an exemplary sample (original in Chinese, with English translation).}
    \label{fig:app_pri_exp2}
\end{figure}

\begin{figure}[!hbt]
    \centering
    \includegraphics[width=0.9\linewidth]{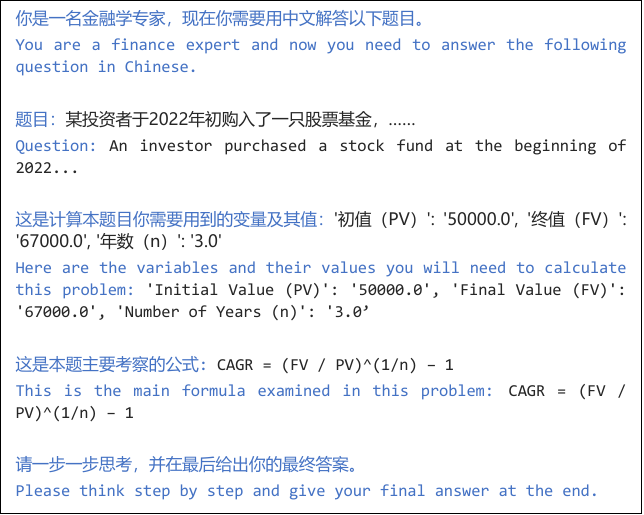}
    \caption{The prompt for experiment 3 and an exemplary sample (original in Chinese, with English translation).}
    \label{fig:app_pri_exp3}
\end{figure}

We manually construct a dataset consisting of 200 samples. Each sample in this dataset includes the question, the financial formula being examined, the mapping between the formula's variables and the specific numerical values, and the ground truth to the question. Figure~\ref{fig:app_pri_exp1}, \ref{fig:app_pri_exp2}, and \ref{fig:app_pri_exp3} show the prompt templates and example samples used in experiments 1, 2, and 3, respectively. In these figures, the content of the prompt template is shown in blue text, while the test samples are shown in black text. The ground truth for this problem is 0.1024 or 10.24\%.

% -----------------------

\section{The Prompt Templates for the Dataset Generation}
\label{sec:app_dataset_generation}
\begin{figure*}[!hbt]
    \centering
    \includegraphics[width=0.9\linewidth]{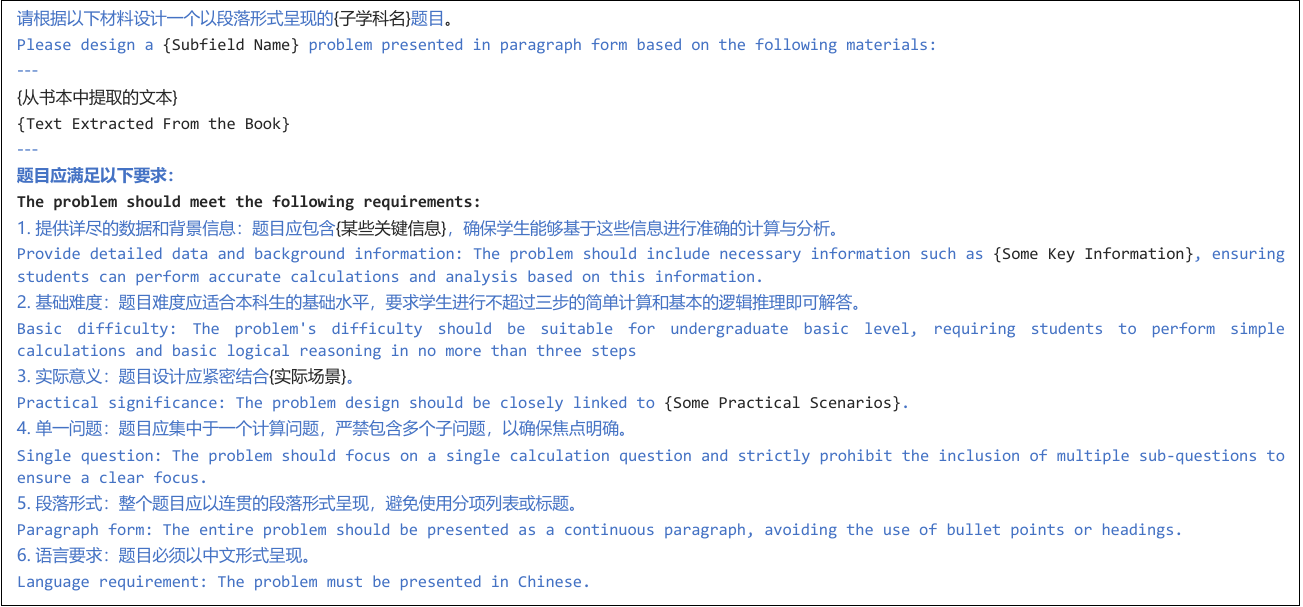}
    \caption{Prompt template for generating questions for a given subfield based on a given piece of corpus (original in Chinese, with English translation).}
    \label{fig:generate_question}
\end{figure*}

\begin{figure}[!hbt]
    \centering
    \includegraphics[width=0.9\linewidth]{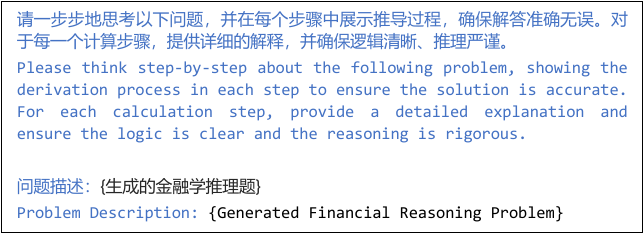}
    \caption{Prompt template that generates an solution to a given question (original in Chinese, with English translation).}
    \label{fig:generate_answer}
\end{figure}

\begin{figure}[!hbt]
    \centering
    \includegraphics[width=0.9\linewidth]{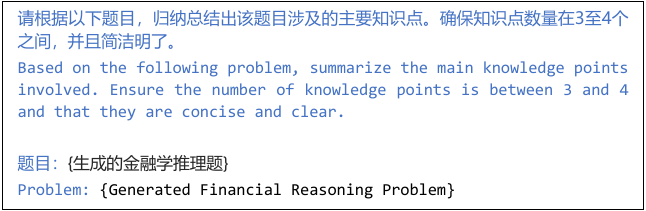}
    \caption{Prompt template for labeling knowledge points for a given question (original in Chinese, with English translation).}
    \label{fig:annotate_knowledge_point}
\end{figure}

\begin{figure*}[!hbt]
    \centering
    \includegraphics[width=0.9\linewidth]{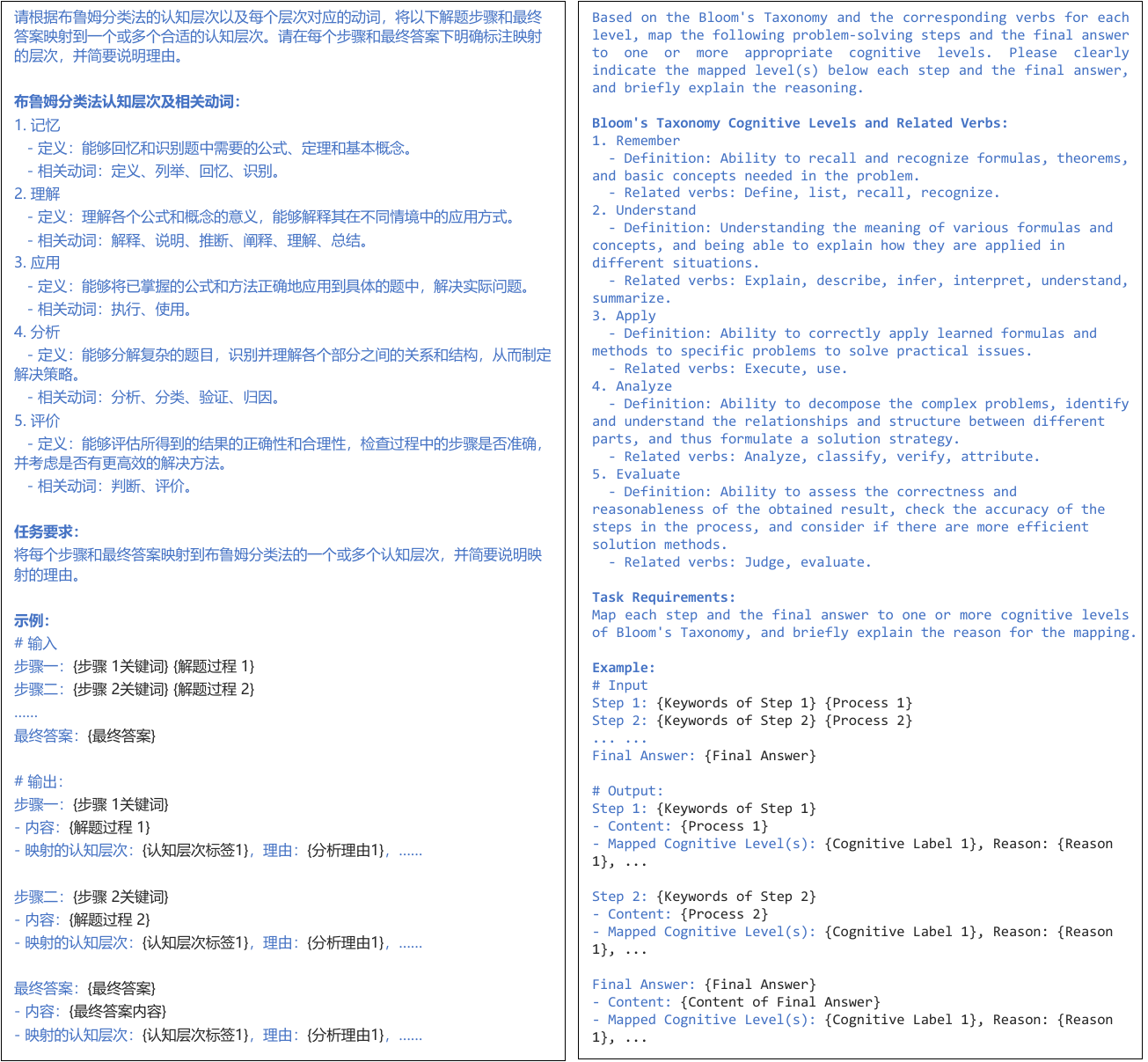}
    \caption{Prompt template for labeling step-level cognitive labels for a given answer (original in Chinese, with English translation).}
    \label{fig:annotate_cognitive_label}
\end{figure*}

Figure ~\ref{fig:generate_question} shows a prompt template for generating questions for a given subfield based on a given piece of corpus. Figure~\ref{fig:generate_answer} shows a prompt template that generates an solution to a given question.
Figures~\ref{fig:annotate_knowledge_point} and \ref{fig:annotate_cognitive_label} show the prompt templates for labeling knowledge points and step-level cognitive abilities, respectively.

% -----------------------
\section{Data Sources for the FinEval-KR Dataset}
\label{sec:app_datasource}
To ensure our benchmark is both authoritative and comprehensive, we constructed the source corpus from nine classic textbooks in modern finance. This selection provides extensive coverage across key subfields, including corporate finance, investments, financial markets, risk management, and monetary policy. These foundational texts supply a rich combination of core theoretical principles and practical case studies, forming a robust basis for evaluating financial knowledge and reasoning.

We processed the corpus using a three-stage pipeline: extraction, cleaning, and standardization. (1) Extraction: We used OCR to convert all text and mathematical equations from the source materials into a machine-readable Markdown format. (2) Cleaning: We then manually curated the extracted content, removing non-essential sections (e.g., prefaces, appendices) and performing quality assurance checks. (3) Standardization: Finally, we transformed the cleaned content into a structured format suitable for automated processing.
This rigorous process ensures the final dataset is of high quality, integrity, and utility.

\paragraph{Corporate Finance}
\begin{itemize}
    \item Selected Textbook: \textit{Corporate Finance} (13th edition, 2021) by Stephen A. Ross, Randolph W. Westerfield, Jeffrey Jaffe, and Bradford D. Jordan.

    \item Rationale: This textbook is widely used in MBA and undergraduate finance courses. It systematically explains core concepts of modern corporate finance, such as arbitrage theory, net present value (NPV), the efficient market hypothesis, agency theory, and the risk-return tradeoff.

    \item Covered Financial Subfields: Corporate financing, capital structure, investment decisions, dividend policy, firm valuation, etc.
    
    \item Role in the Benchmark Dataset: Provides a solid theoretical foundation and abundant practical examples for reasoning and computational questions in corporate finance.
\end{itemize}

\paragraph{Investments}
\begin{itemize}
    \item Selected Textbook: \textit{Investments} (13th edition, 2023) by Zvi Bodie, Alex Kane, and Alan J. Marcus.

    \item Rationale: This book deeply explores securities market efficiency, risk-return relationships, and asset allocation strategies. Its content is highly aligned with the CFA (Chartered Financial Analyst) exam syllabus.

    \item Covered Financial Subfields: Securities markets, asset pricing, portfolio theory, behavioral finance, derivatives, etc.

    \item Role in the Benchmark Dataset: Offers authoritative theoretical support and practical guidance for reasoning and computational questions in investments.
\end{itemize}

\paragraph{Financial Institutions and Markets}
\begin{itemize}
    \item Selected Textbook: \textit{Financial Markets \& Institutions} (13th edition, 2020) by Jeff Madura.

    \item Rationale: This book comprehensively analyzes the operational mechanisms and regulatory frameworks of financial institutions like commercial and investment banks. It also provides empirical and case analyses on contemporary hot topics such as stock valuation and market microstructure.

    \item Covered Financial Subfields: Financial institutions, financial markets, central banking, monetary policy, market regulation, etc.

    \item Role in the Benchmark Dataset: Supplies a systematic theoretical framework and practical examples for reasoning and computational questions concerning financial institutions and markets.
\end{itemize}

\paragraph{Money and Banking}
\begin{itemize}
    \item Selected Textbook: \textit{The Economics of Money, Banking, and Financial Markets} (13th edition, 2021) by Frederic S. Mishkin.

    \item Rationale: This book offers an in-depth analysis, from both theoretical and empirical perspectives, of money demand and supply, commercial banking operations and regulation, and the interaction mechanisms between monetary policy tools and financial markets.

    \item Covered Financial Subfields: Monetary theory, banking systems, monetary policy, financial markets, etc.

    \item Role in the Benchmark Dataset: Delivers in-depth theoretical analysis and empirical support for reasoning and computational questions in money and banking.
\end{itemize}

\paragraph{International Finance}
\begin{itemize}
    \item Selected Textbook: \textit{International Financial Management} (6th edition, 2023) by Jeff Madura and Roland Fox.

    \item Rationale: In the context of globalization, this textbook discusses cross-border capital flows, exchange rate volatility, and risk management strategies. It uses numerous case studies to examine practical operations in the international financial environment.

    \item Covered Financial Subfields: International capital flows, exchange rate theory, foreign exchange markets, international investment, multinational corporate financial management, etc.

    \item Role in the Benchmark Dataset: Provides a global perspective and practical examples for reasoning and computational questions in international finance.
\end{itemize}

\paragraph{Financial Risk Management}
\begin{itemize}
    \item Selected Textbook: \textit{Risk Management and Financial Institutions} (6th edition, 2022) by John C. Hull.

    \item Rationale: This book comprehensively reviews methods for measuring and hedging market risk, credit risk, and operational risk. It places particular emphasis on the application of financial derivatives in risk management.

    \item Covered Financial Subfields: Risk management, financial derivatives, financial institution regulation, risk measurement and hedging, etc.

    \item Role in the Benchmark Dataset: Offers a systematic risk analysis framework and practical guidance for reasoning and computational questions in financial risk management.
\end{itemize}

\paragraph{Fixed Income Securities}
\begin{itemize}
    \item Selected Textbook: \textit{Fixed Income Securities: Tools for Today's Markets} (4th edition, 2022) by Bruce Tuckman and Angel Serrat.

    \item Rationale: This book provides detailed discussions on the pricing principles and trading strategies for fixed income products such as government bonds, interest rate swaps, and credit default swaps.

    \item Covered Financial Subfields: Fixed income securities, bond pricing, interest rate derivatives, credit risk, etc.

    \item Role in the Benchmark Dataset: Provides authoritative pricing models and practical examples for reasoning and computational questions related to fixed income securities.
\end{itemize}

\paragraph{Financial Engineering and Derivatives}
\begin{itemize}
    \item Selected Textbook: \textit{Options, Futures, and Other Derivatives} (10th edition, 2018) by John C. Hull and Basu Sankarshan.

    \item Rationale: This textbook comprehensively covers core topics in financial engineering, including option pricing models, futures contract structures, and the pricing of interest rate and credit derivatives.

    \item Covered Financial Subfields: Derivatives markets, option pricing, futures contracts, interest rate derivatives, credit derivatives, etc.

    \item Role in the Benchmark Dataset: Offers in-depth theoretical analysis and practical guidance for reasoning and computational questions in financial engineering and derivatives.
\end{itemize}

\paragraph{Monetary Theory and Policy}
\begin{itemize}
    \item Selected Textbook: \textit{Monetary Theory and Policy} (4th edition, 2017) by Carl E. Walsh.

    \item Rationale: This book systematically explains the framework of modern monetary theory, focusing on the transmission mechanisms of various monetary policy tools and their effectiveness in low-interest-rate environments.

    \item Covered Financial Subfields: Monetary theory, monetary policy, macroeconomic models, policy transmission mechanisms, etc.

    \item Role in the Benchmark Dataset: Provides a macroeconomic perspective and policy analysis framework for reasoning and computational questions in monetary theory and policy.
\end{itemize}

In summary, these nine textbooks are not only authoritative and reliable but also closely aligned with current academic frontiers. They lay a comprehensive and in-depth academic foundation for the financial reasoning and computation benchmark dataset constructed in this study. This ensures that the test questions possess both professional depth and practical relevance.

\section{Validation in Question and Answer Generation}
\label{sec:app_validation}

For both the question and answer generation phases, we adopted a three-stage verification process. The verification focus for each stage is detailed in Table~\ref{tab:validation_question} and Table~\ref{tab:validation_answer}, respectively.

\begin{table*}[!t]
\centering
\scalebox{0.65}{
\begin{tabularx}{1.4\linewidth}{@{}lXX@{}}
\toprule
Stage  & Aspect      & Criteria       \\ \midrule
\multirow{3}{*}{Logical Validation}       & Clarity and Completeness Check    & (1) Is the question description clear and unambiguous? (2) Is there any misuse of terminology? (3) Are all necessary conditions and data for calculation provided? \\ \cmidrule(l){2-3} 
                                        & Plausibility Check    & (1) Are numerical values (e.g., interest rates, returns, prices) within a plausible range? (2) Is the scenario self-contradictory or unrealistic?\\ \cmidrule(l){2-3} 
                                        & Solvability Check     & Assuming the data is complete and plausible, does the question have a deterministic solution that can be calculated using financial models?     \\ \midrule
Consistency Assessment                  & Relevance Check       & Do the core concepts in the question align with the title or keywords of the corresponding textbook chapter?         \\ \midrule
Final Validation                        & Samples that fail any of the above checks are marked as "unqualified" and removed from the final dataset. &    \\ \bottomrule
\end{tabularx}
}
\caption{Validation in question generation stage.}
\label{tab:validation_question}
\end{table*}

\begin{table*}[!t]
\centering
\scalebox{0.65}{
\begin{tabularx}{1.4\linewidth}{@{}lXX@{}}
\toprule
Stages & Aspect   & Criteria     \\ \midrule
\multirow{3}{*}{Logical Validation}    & Formula/Model Selection Check    & (1) Is the selected formula a standard method for this type of problem? (2) Does the variable required by the question match the variable solved for in the answer? \\  \cmidrule(l){2-3} 
                                      & Parameter Substitution Check       & Do the numerical values from the question correctly correspond to the variables in the formula during calculation?   \\ \midrule
Consistency Assessment                & Calculation Validation              & Execute the calculation code output by OpenAI o1 to verify the answer's correctness.    \\ \midrule
Final Validation                      & Samples that fail any of the above checks are marked as "unqualified" and removed from the final dataset. &     \\ \bottomrule
\end{tabularx}
}
\caption{Validation in answer generation stage.}
\label{tab:validation_answer}
\end{table*}

% -----------------------
\section{Statistical Characterization of the FinEval-KR Dataset}
\label{sec:appendix_stat}
\begin{figure*}
    \centering
    \includegraphics[width=0.9\linewidth]{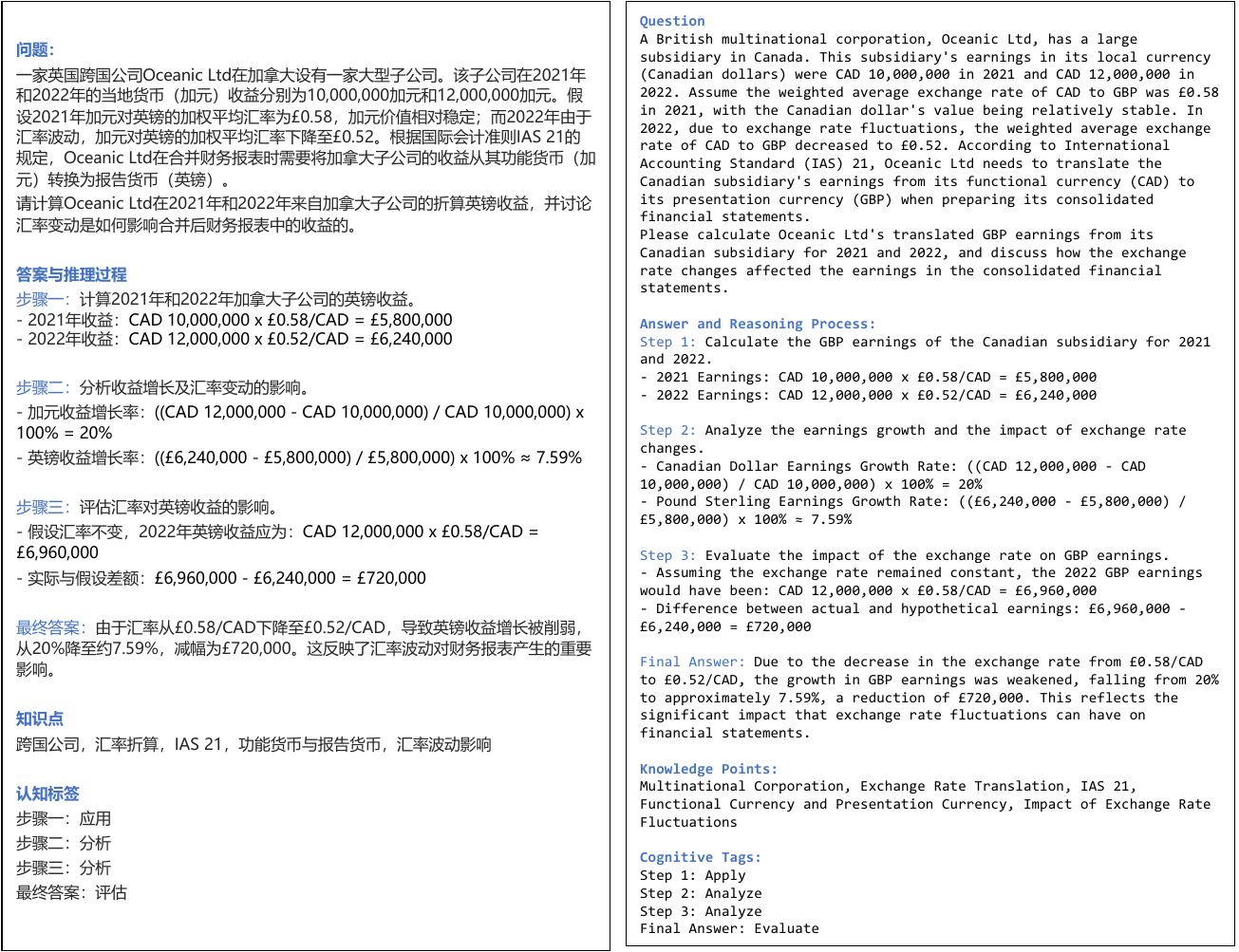}
    \caption{A complete sample from FinEval-KR dataset (original in Chinese, with English translation).}
    \label{fig:full_example}
\end{figure*}

\begin{figure}[!bthp]
    \centering
    \includegraphics[width=0.9\linewidth]{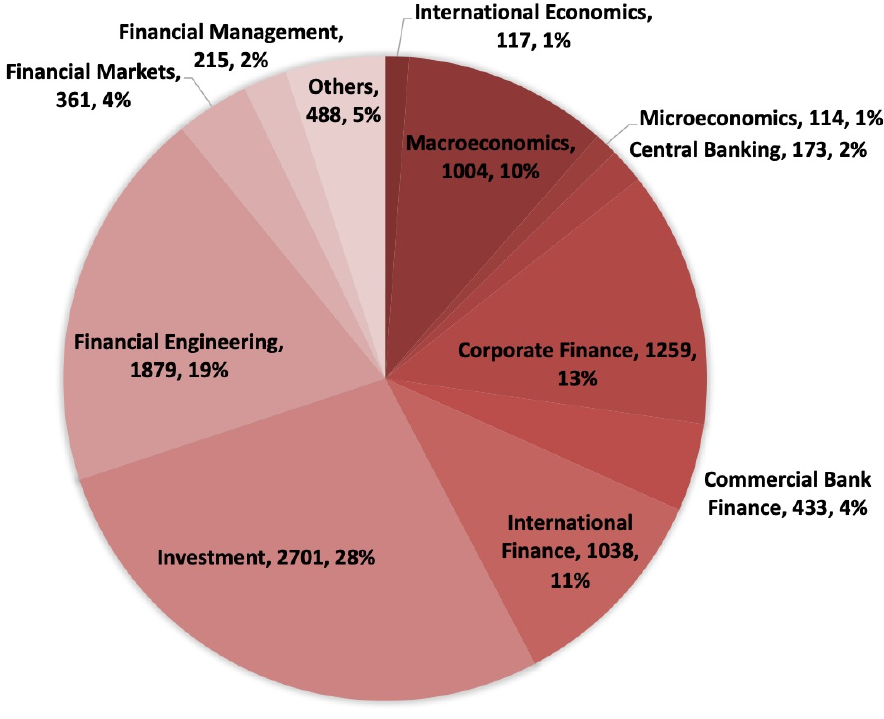}
    \caption{The number of samples in each subdiscipline in the FinEval-KR dataset and their percentage, and ``others'' in the pie chart includes: econometrics, public finance, insurance, monetary economics, managerial accounting, intermediate financial accounting, corporate strategy and risk management, auditing, cost accounting, taxation and advanced financial accounting.}
    \label{fig:dataset_stat}
\end{figure}

\begin{figure*}[!bt]
    \centering
    \includegraphics[width=0.85\linewidth]{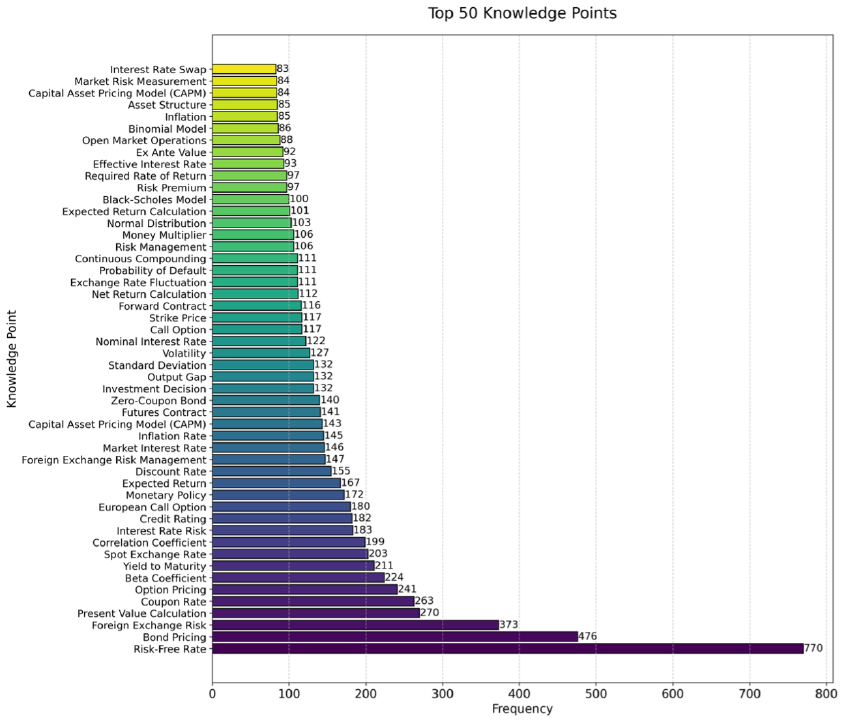}
    \caption{Top 50 knowledge points in the dataset.}
    \label{fig:top50}
\end{figure*}

\begin{figure*}[!bt]
    \centering
    \includegraphics[width=0.8\linewidth]{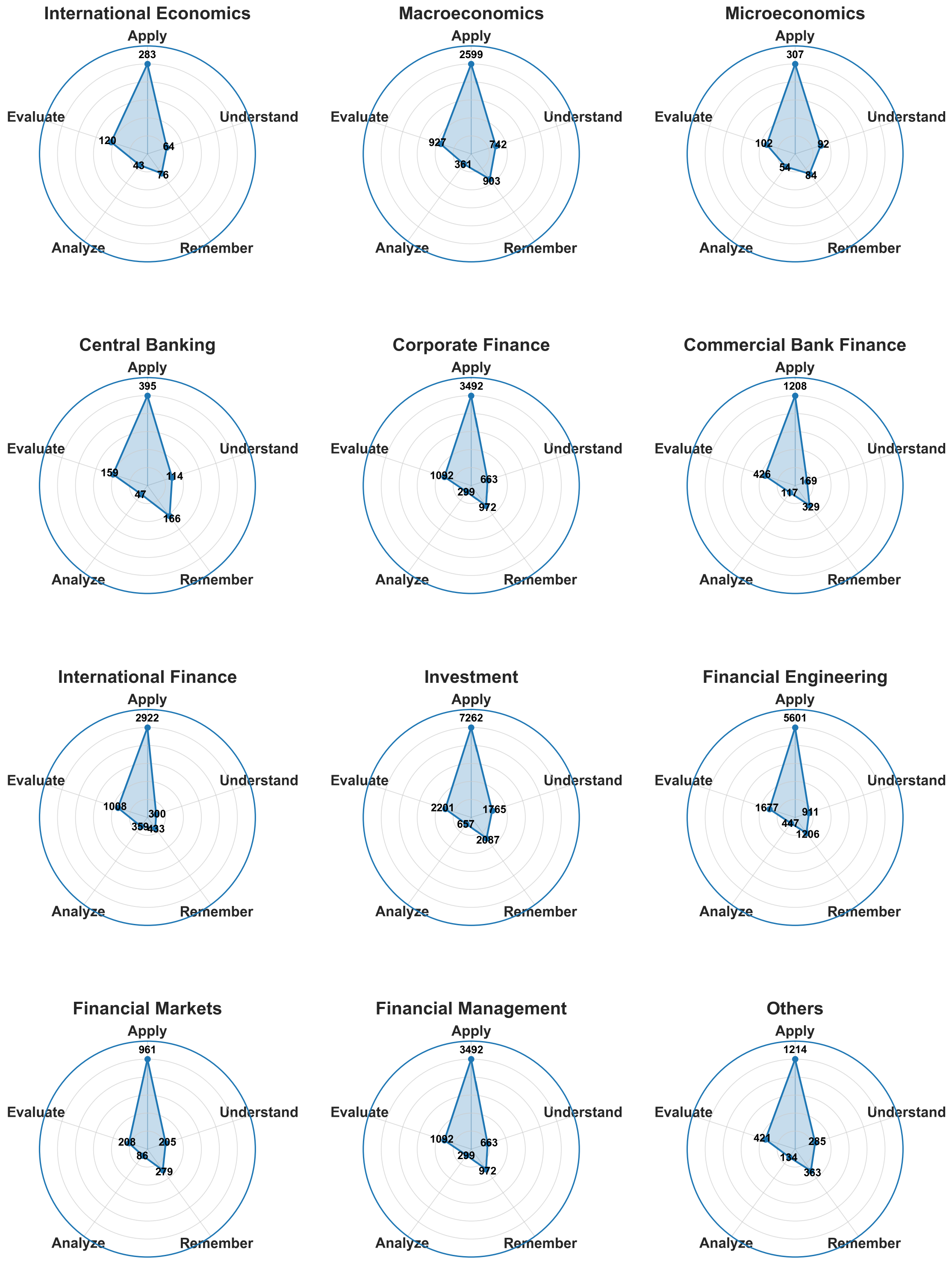}
    \caption{The distribution of cognitive labels in each subdiscipline.}
    \label{fig:radar_charts}
\end{figure*}

A complete sample from the constructed dataset is shown in Figure~\ref{fig:full_example}.
The sample size and its distribution for each subdiscipline in FinEval-KR Dataset is shown in Figure~\ref{fig:dataset_stat}, where the subdiscipline categorization methodology refers to~\citet{zhang2023fineval}.
The top-50 knowledge points in the dataset are shown in Figure~\ref{fig:top50}.
In each subdiscipline, the distribution of cognitive labels is shown in Figure~\ref{fig:radar_charts}.

% -----------------------

\section{Prompt Template Adopted by the FinEval-KR Evaluation Framework}
\label{sec:app_framework}
\begin{figure*}[!hbtp]
    \centering
    \includegraphics[width=0.9\linewidth]{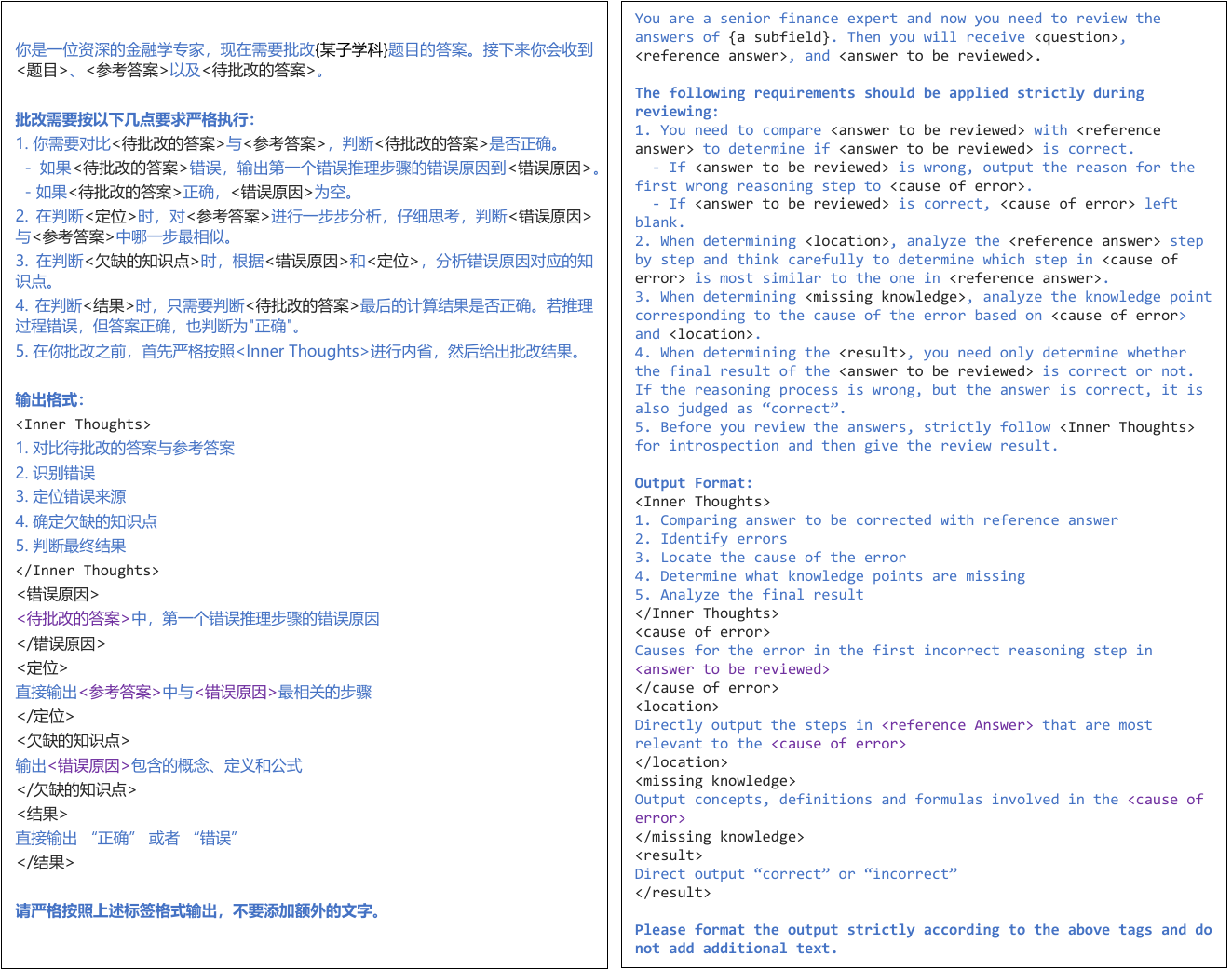}
    \caption{Prompt templates for the judge model (original in Chinese, with English translation).}
    \label{fig:judge}
\end{figure*}

\begin{figure}[!hbt]
    \centering
    \includegraphics[width=0.9\linewidth]{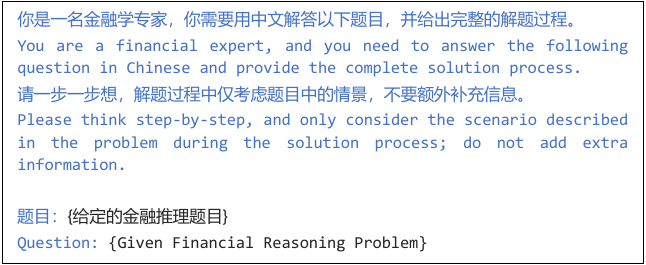}
    \caption{Prompt template for Stage 1 where the evaluated model answers the questions in a free-form format (original in Chinese, with English translation).}
    \label{fig:question_answering}
\end{figure}

\begin{figure}[!hbt]
    \centering
    \includegraphics[width=0.9\linewidth]{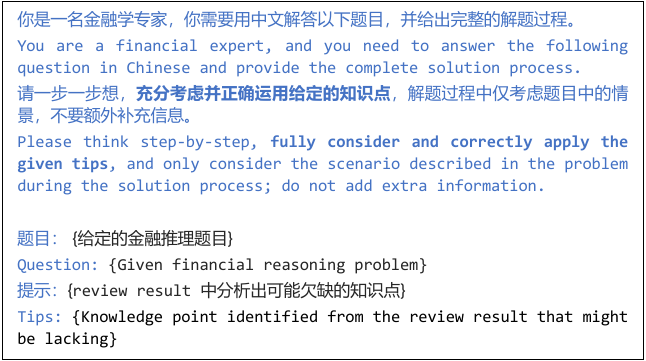}
    \caption{Prompt template for Stage 2 where the evaluated model re-answer the question with knowledge point augmented (original in Chinese, with English translation).}
    \label{fig:re-answer}
\end{figure}

Figure~\ref{fig:question_answering} shows the prompt template used in Stage 1 where the evaluated model answers the questions in a free-form format.
Figure~\ref{fig:re-answer} shows the prompt template used in Stage 2, in which the evaluated model re-answer the question with knowledge point augmented.
% Note that we have made every effort to ensure the consistency of the prompt structure used in Stage 1 and Stage 2 to avoid changes in reasoning performance due to structural variations. During the prompt design, we conduct multiple rounds of trials and adjustments, ultimately ensuring that changes in the model's responses are solely related to the injected knowledge points.
Figure~\ref{fig:judge} shows the prompt templates for the judge model to generate review results and augmented review results based on the reference answers in Stage 1 and 2.

% -----------------------

\section{Bias Challenges in Judge Model}
\label{sec:judge_bias}
Previous research has shown that using LLMs as judges to evaluate the output of other models inevitably introduces certain evaluation biases, which may lead to unfair comparison results. Therefore, this section will discuss the main evaluation biases discovered during the experimental process of this study and their corresponding mitigation methods.
\begin{itemize}
    \item \textit{Style Bias}: This bias refers to the tendency of the judge LLM to give higher scores to content with a more appealing text style (e.g., clear structure, moderate length), even if the answers contain reasoning errors~\cite{wu2025style}. To reduce the impact of this type of bias, we did not restrict the output format of the models being evaluated in Stage 1 and Stage 2, encouraging them to reason freely. Subsequently, we used methods such as regular expressions to unify the original output of each model into a format consistent with the reference answer. This processing method effectively reduced evaluation biases caused by differences in text style.

    \item \textit{Cognitive Bias}: This bias refers to the self-bias that LLMs may exhibit during the evaluation, i.e., a tendency to give higher scores to content they generated themselves, thereby affecting the fairness of the evaluation~\cite{koo2024benchmarking}. To avoid this type of cognitive bias, we excluded OpenAI o1 from the scope of evaluated models in our experiment, as it had already been used in the dataset construction and validation. Furthermore, during the preliminary experiment, we tested whether Qwen2.5-72B\_Instruct, used as the judging model, exhibited significant cognitive bias in the root cause localization and knowledge gap identification tasks. The experimental results showed that this model did not exhibit a significant self-bias tendency in the aforementioned two tasks. We believe this may be because these two tasks are more specific and objective compared to result comparison tasks without sub-task decomposition, and are further aided by the model introspection prompting shown in Figure~\ref{fig:judge}, which helps enhance the objectivity of Qwen2.5-72B\_Instruct during the evaluation.

    \item To prevent the judge model from the disturbance of ``simple deception''~\cite{thakur2024judging}, we filter out meaningless content in the generated answers, such as isolated affirmative words like ``yes'' or ``of course'', ensuring that the evaluation focuses on the substantive reasoning process rather than superficial linguistic features.
\end{itemize}

\section{Details of Experiment}
\label{sec:app_models}
\paragraph{Evaluated Open-source Models}
We select several popular LLMs, including DeepSeek-R1, DeepSeek-V3, and QwQ-32B-preview. To study the effect of model size on performance, we also include smaller models, such as Qwen2.5-14B\_Instruct. In total, this study includes four open-source LLMs.

\paragraph{Evaluated Close-source Models}
We include a selection of prominent models. This selection covers the latest reasoning models, such as Claude-3.7-sonnet, Gemini-2.5-flash, Gemini-2.5-pro, o3-mini, and o1-mini. We also include the current top non-reasoning models: GPT-4.1, GPT-4o, and Qwen-max-latest. Furthermore, we add some models released in 2024 and 2023, including Moonshot-v1-128k, Doubao-pro-32k, Ernie-Bot-4.0, and GPT-3.5-turbo.

\paragraph{Financial LLMs}
Furthermore, we specifically select two financial reasoning LLMs for evaluation. The first is Xuanyuan-FinX1-preview from Duxiaoman AI-Lab, a Chinese financial dialogue and reasoning model designed specifically for the financial domain. It is also the first o1-like model in the financial industry. The second is Fin-R1, a financial reasoning LLM jointly developed by Shanghai University of Finance and Economics and StepFun Technology. This model is trained on Qwen2.5-7B\_Instruct and designed for complex financial reasoning tasks, balancing high performance with low deployment cost.

\paragraph{Implement}
During evaluation, all closed-source models are accessed through the official APIs provided by their respective developers. In contrast, open-source models are accessed using the service provided by either Bailian\footnote{\url{https://bailian.console.aliyun.com}}or ModelScope\footnote{\url{https://www.modelscope.cn}}.

\begin{table}[!t]
\centering
\scalebox{0.8}{
\begin{tabular}{@{}cl@{}}
\toprule
Model                             & Version    \\ \midrule
Qwen2.5-14B\_Instruct              & 2024-09-19 \\
QwQ-32B-preview                   & 2025-03-06 \\
DeepSeek-V3                 & 2025-03-24 \\
DeepSeek-R1                 & 2025-01-20 \\
Doubao-pro-32k                    & 2024-06-15 \\
Moonshot-v1-128k                  & 2024-01-31 \\
Ernie-Bot-4.0                     & 2023-11-17 \\
Qwen-max-latest                   & 2025-01-25 \\
GPT-3.5-turbo                     & 2024-01-25 \\
GPT-4o                            & 2024-11-20 \\
GPT-4.1                           & 2025-04-14 \\
Gemini-2.5-pro                    & 2025-05-06 \\
Claude-3.7-sonnet                 & 2025-02-19 \\
o1-mini                       & 2024-09-12 \\
o3-mini                       & 2025-01-31 \\
Gemini-2.5-flash                  & 2025-04-17 \\
Xuanyuan-FinX1-preview            & 2024-12-27 \\
Fin-R1                            & 2025-03-22 \\ \bottomrule
\end{tabular}
}
\caption{Version of the model being evaluated.}
\label{tab:model_version}
\end{table}

% -----------------------
\section{Discussion of Experimental Results}
\label{sec:app_experiment}
We evaluate 18 LLMs listed in Appendix~\ref{sec:app_models} using our proposed financial reasoning dataset. Table~\ref{tab:all_results} presents the complete evaluation results across several metrics: Knowledge Score (KS), Reasoning Score (RS), Cognitive Scores (CS$_1$ to CS$_5$), and Task Accuracy (Acc). Results in the table are from three runs of each model. The CS$_1$ to CS$_5$ correspond to remembering, understanding, applying, analyzing, and evaluating in Bloom's taxonomy, respectively. 

For all subsequent analyses, our focus is on performance tiers instead of absolute scores, which helps alleviate assessment errors caused by the systematic bias and randomness of the judge model. Additionally, we set the distribution of models across the tiers to 3:4:5:6.

\begin{table*}[t]
\centering
\scalebox{0.52}{
\begin{tabular}{@{}ccccccccccccccccc@{}}
\toprule
Model/Metrics          & Acc             & \cellcolor[HTML]{EFEFEF}Acc.std & KS              & \cellcolor[HTML]{EFEFEF}KS.std & RS              & \cellcolor[HTML]{EFEFEF}RS.std & \makecell[c]{CS$_1$\\(remember)} & \cellcolor[HTML]{EFEFEF}CS$_1$.std & \makecell[c]{CS$_2$\\(understand)} & \cellcolor[HTML]{EFEFEF}CS$_2$.std & \makecell[c]{CS$_3$\\(apply)}   & \cellcolor[HTML]{EFEFEF}CS$_3$.std & \makecell[c]{CS$_4$\\(analyze)} & \cellcolor[HTML]{EFEFEF}CS$_4$.std & \makecell[c]{CS$_5$\\(evaluate)} & \cellcolor[HTML]{EFEFEF}CS$_5$.std \\ \midrule
\multicolumn{17}{c}{Open-source lightweight LLMs without reasoning}                                                                                                                                                                                                                                                                                                                                                                                                 \\ \midrule
Qwen2.5-14B\_Instruct  & 0.5473          & \cellcolor[HTML]{EFEFEF}0.0006  & 0.8490          & \cellcolor[HTML]{EFEFEF}0.0010 & 0.6863          & \cellcolor[HTML]{EFEFEF}0.0038 & 0.6547           & \cellcolor[HTML]{EFEFEF}0.0015    & 0.6603             & \cellcolor[HTML]{EFEFEF}0.0023    & 0.3893          & \cellcolor[HTML]{EFEFEF}0.0064    & 0.6863          & \cellcolor[HTML]{EFEFEF}0.0038    & 0.6820           & \cellcolor[HTML]{EFEFEF}0.0046    \\ \midrule
\multicolumn{17}{c}{Open-source lightweight LLMs with reasoning}                                                                                                                                                                                                                                                                                                                                                                                                    \\ \midrule
QwQ-32B-preview        & 0.7380          & \cellcolor[HTML]{EFEFEF}0.0061  & 0.9073          & \cellcolor[HTML]{EFEFEF}0.0057 & 0.8627          & \cellcolor[HTML]{EFEFEF}0.0136 & 0.8450           & \cellcolor[HTML]{EFEFEF}0.0141    & 0.8503             & \cellcolor[HTML]{EFEFEF}0.0143    & 0.6987          & \cellcolor[HTML]{EFEFEF}0.0267    & 0.8510          & \cellcolor[HTML]{EFEFEF}0.0075    & 0.8597           & \cellcolor[HTML]{EFEFEF}0.0136    \\ \midrule
\multicolumn{17}{c}{Open-source LLMs without reasoning}                                                                                                                                                                                                                                                                                                                                                                                                             \\ \midrule
DeepSeek-v3            & 0.8270          & \cellcolor[HTML]{EFEFEF}0.0062  & 0.9427          & \cellcolor[HTML]{EFEFEF}0.0050 & 0.9077          & \cellcolor[HTML]{EFEFEF}0.0050 & 0.8963           & \cellcolor[HTML]{EFEFEF}0.0059    & 0.8993             & \cellcolor[HTML]{EFEFEF}0.0059    & 0.7963          & \cellcolor[HTML]{EFEFEF}0.0125    & 0.8943          & \cellcolor[HTML]{EFEFEF}0.0075    & 0.9057           & \cellcolor[HTML]{EFEFEF}0.0057    \\ \midrule
\multicolumn{17}{c}{Open-source LLMs with reasoning}                                                                                                                                                                                                                                                                                                                                                                                                                \\ \midrule
DeepSeek-R1            & 0.8700          & \cellcolor[HTML]{EFEFEF}0.0165  & 0.9517          & \cellcolor[HTML]{EFEFEF}0.0171 & \textbf{0.9347} & \cellcolor[HTML]{EFEFEF}0.0153 & \textbf{0.9377}  & \cellcolor[HTML]{EFEFEF}0.0186    & \textbf{0.9397}    & \cellcolor[HTML]{EFEFEF}0.0179    & \textbf{0.8810} & \cellcolor[HTML]{EFEFEF}0.0358    & \textbf{0.9380} & \cellcolor[HTML]{EFEFEF}0.0190    & \textbf{0.9433}  & \cellcolor[HTML]{EFEFEF}0.0158    \\ \midrule
\multicolumn{17}{c}{Closed-source LLMs without reasoning}                                                                                                                                                                                                                                                                                                                                                                                                           \\ \midrule
Doubao-pro-32k         & 0.7825          & \cellcolor[HTML]{EFEFEF}0.0007  & 0.9195          & \cellcolor[HTML]{EFEFEF}0.0007 & 0.8750         & \cellcolor[HTML]{EFEFEF}0.0057 & 0.8560           & \cellcolor[HTML]{EFEFEF}0.0071    & 0.8600             & \cellcolor[HTML]{EFEFEF}0.0085    & 0.7340          & \cellcolor[HTML]{EFEFEF}0.0113    & 0.8565          & \cellcolor[HTML]{EFEFEF}0.0064    & 0.8720           & \cellcolor[HTML]{EFEFEF}0.0057    \\
Moonshot-v1-128k       & 0.4533          & \cellcolor[HTML]{EFEFEF}0.0015  & 0.8340          & \cellcolor[HTML]{EFEFEF}0.0061 & 0.6020          & \cellcolor[HTML]{EFEFEF}0.0082 & 0.5620           & \cellcolor[HTML]{EFEFEF}0.0108    & 0.5670             & \cellcolor[HTML]{EFEFEF}0.0087    & 0.2763          & \cellcolor[HTML]{EFEFEF}0.0064    & 0.5653          & \cellcolor[HTML]{EFEFEF}0.0074    & 0.5973           & \cellcolor[HTML]{EFEFEF}0.0074    \\
Ernie-bot-4.0          & 0.5733          & \cellcolor[HTML]{EFEFEF}0.0025  & 0.8627          & \cellcolor[HTML]{EFEFEF}0.0081 & 0.7053          & \cellcolor[HTML]{EFEFEF}0.0091 & 0.6680           & \cellcolor[HTML]{EFEFEF}0.0089    & 0.6753             & \cellcolor[HTML]{EFEFEF}0.0091    & 0.4383          & \cellcolor[HTML]{EFEFEF}0.0146    & 0.6847          & \cellcolor[HTML]{EFEFEF}0.0074    & 0.6927           & \cellcolor[HTML]{EFEFEF}0.0251    \\
Qwen-max-latest        & 0.6467          & \cellcolor[HTML]{EFEFEF}0.0015  & 0.8797          & \cellcolor[HTML]{EFEFEF}0.0050 & 0.7733          & \cellcolor[HTML]{EFEFEF}0.0042 & 0.7507           & \cellcolor[HTML]{EFEFEF}0.0050    & 0.7547             & \cellcolor[HTML]{EFEFEF}0.0057    & 0.5340          & \cellcolor[HTML]{EFEFEF}0.0040    & 0.7440          & \cellcolor[HTML]{EFEFEF}0.0026    & 0.7703           & \cellcolor[HTML]{EFEFEF}0.0042    \\
GPT-3.5-turbo          & 0.2830          & \cellcolor[HTML]{EFEFEF}0.0040  & 0.7527          & \cellcolor[HTML]{EFEFEF}0.0038 & 0.3973          & \cellcolor[HTML]{EFEFEF}0.0040 & 0.3527           & \cellcolor[HTML]{EFEFEF}0.0021    & 0.3603             & \cellcolor[HTML]{EFEFEF}0.0021    & 0.0900          & \cellcolor[HTML]{EFEFEF}0.0036    & 0.3893          & \cellcolor[HTML]{EFEFEF}0.0081    & 0.3970           & \cellcolor[HTML]{EFEFEF}0.0036    \\
GPT-4o                 & 0.6853          & \cellcolor[HTML]{EFEFEF}0.0142  & 0.9020          & \cellcolor[HTML]{EFEFEF}0.0159 & 0.8067          & \cellcolor[HTML]{EFEFEF}0.0080 & 0.7847           & \cellcolor[HTML]{EFEFEF}0.0081    & 0.7890             & \cellcolor[HTML]{EFEFEF}0.0090    & 0.5930          & \cellcolor[HTML]{EFEFEF}0.0145    & 0.7870          & \cellcolor[HTML]{EFEFEF}0.0110    & 0.8030           & \cellcolor[HTML]{EFEFEF}0.0085    \\
GPT-4.1                & 0.8263          & \cellcolor[HTML]{EFEFEF}0.0025  & 0.9520          & \cellcolor[HTML]{EFEFEF}0.0040 & 0.9063          & \cellcolor[HTML]{EFEFEF}0.0015 & 0.8957           & \cellcolor[HTML]{EFEFEF}0.0021    & 0.8977             & \cellcolor[HTML]{EFEFEF}0.0025    & 0.7890          & \cellcolor[HTML]{EFEFEF}0.0036    & 0.8927          & \cellcolor[HTML]{EFEFEF}0.0015    & 0.9050           & \cellcolor[HTML]{EFEFEF}0.0017    \\ \midrule
\multicolumn{17}{c}{Closed-source LLMs with reasoning}                                                                                                                                                                                                                                                                                                                                                                                                              \\ \midrule
o1-mini                & 0.7503          & \cellcolor[HTML]{EFEFEF}0.0031  & 0.8997          & \cellcolor[HTML]{EFEFEF}0.0076 & 0.8453          & \cellcolor[HTML]{EFEFEF}0.0067 & 0.8340           & \cellcolor[HTML]{EFEFEF}0.0066    & 0.8363             & \cellcolor[HTML]{EFEFEF}0.0031    & 0.6983          & \cellcolor[HTML]{EFEFEF}0.0081    & 0.8477          & \cellcolor[HTML]{EFEFEF}0.0070    & 0.8450           & \cellcolor[HTML]{EFEFEF}0.0060    \\
o3-mini                & 0.8207          & \cellcolor[HTML]{EFEFEF}0.0095  & 0.9260          & \cellcolor[HTML]{EFEFEF}0.0106 & 0.9070          & \cellcolor[HTML]{EFEFEF}0.0052 & 0.9047           & \cellcolor[HTML]{EFEFEF}0.0102    & 0.9073             & \cellcolor[HTML]{EFEFEF}0.0099    & 0.8127          & \cellcolor[HTML]{EFEFEF}0.0110    & 0.9023          & \cellcolor[HTML]{EFEFEF}0.0107    & 0.9120           & \cellcolor[HTML]{EFEFEF}0.0113    \\
Gemini-2.5-pro         & \textbf{0.8750} & \cellcolor[HTML]{EFEFEF}0.0079  & \textbf{0.9627} & \cellcolor[HTML]{EFEFEF}0.0134 & 0.9233          & \cellcolor[HTML]{EFEFEF}0.0238 & 0.9123           & \cellcolor[HTML]{EFEFEF}0.0272    & 0.9163             & \cellcolor[HTML]{EFEFEF}0.0290    & 0.8403          & \cellcolor[HTML]{EFEFEF}0.0291    & 0.9050          & \cellcolor[HTML]{EFEFEF}0.0260    & 0.9120           & \cellcolor[HTML]{EFEFEF}0.0243    \\
Gemini-2.5-flash       & 0.8440          & \cellcolor[HTML]{EFEFEF}0.0061  & 0.9540          & \cellcolor[HTML]{EFEFEF}0.0020 & 0.9203          & \cellcolor[HTML]{EFEFEF}0.0091 & 0.9103           & \cellcolor[HTML]{EFEFEF}0.0100    & 0.9133             & \cellcolor[HTML]{EFEFEF}0.0108    & 0.8307          & \cellcolor[HTML]{EFEFEF}0.0061    & 0.9100          & \cellcolor[HTML]{EFEFEF}0.0104    & 0.9177           & \cellcolor[HTML]{EFEFEF}0.0110    \\
Claude-3.7-sonnet      & 0.7923          & \cellcolor[HTML]{EFEFEF}0.0040  & 0.9390          & \cellcolor[HTML]{EFEFEF}0.0030 & 0.8823          & \cellcolor[HTML]{EFEFEF}0.0086 & 0.8663           & \cellcolor[HTML]{EFEFEF}0.0100    & 0.8703             & \cellcolor[HTML]{EFEFEF}0.0096    & 0.7433          & \cellcolor[HTML]{EFEFEF}0.0120    & 0.8653          & \cellcolor[HTML]{EFEFEF}0.0093    & 0.8803           & \cellcolor[HTML]{EFEFEF}0.0086    \\ \midrule
\multicolumn{17}{c}{Financial LLMs with reasoning}                                                                                                                                                                                                                                                                                                                                                                                                                  \\ \midrule
Xuanyuan-FinX1-preview & 0.5890          & \cellcolor[HTML]{EFEFEF}0.0026  & 0.8687          & \cellcolor[HTML]{EFEFEF}0.0032 & 0.7323          & \cellcolor[HTML]{EFEFEF}0.0042 & 0.7063           & \cellcolor[HTML]{EFEFEF}0.0032    & 0.7130             & \cellcolor[HTML]{EFEFEF}0.0044    & 0.4610          & \cellcolor[HTML]{EFEFEF}0.0066    & 0.7323          & \cellcolor[HTML]{EFEFEF}0.0042    & 0.7300           & \cellcolor[HTML]{EFEFEF}0.0035    \\
Fin-R1-7B              & 0.4153          & \cellcolor[HTML]{EFEFEF}0.0031  & 0.7510          & \cellcolor[HTML]{EFEFEF}0.0346 & 0.5570          & \cellcolor[HTML]{EFEFEF}0.0040 & 0.5190           & \cellcolor[HTML]{EFEFEF}0.0046    & 0.5277             & \cellcolor[HTML]{EFEFEF}0.0065    & 0.2170          & \cellcolor[HTML]{EFEFEF}0.0036    & 0.5570          & \cellcolor[HTML]{EFEFEF}0.0040    & 0.5527           & \cellcolor[HTML]{EFEFEF}0.0045    \\ \bottomrule
\end{tabular}
}
\caption{The complete evaluation results across several metrics: Knowledge Score (KS), Reasoning Score (RS), Cognitive Scores (CS$_1$ to CS$_5$), and Task Accuracy (Acc).}
\label{tab:all_results}
\end{table*}

\subsection{Analysis of Knowledge Score}

\begin{figure}[!t]
    \centering
    \includegraphics[width=0.9\linewidth]{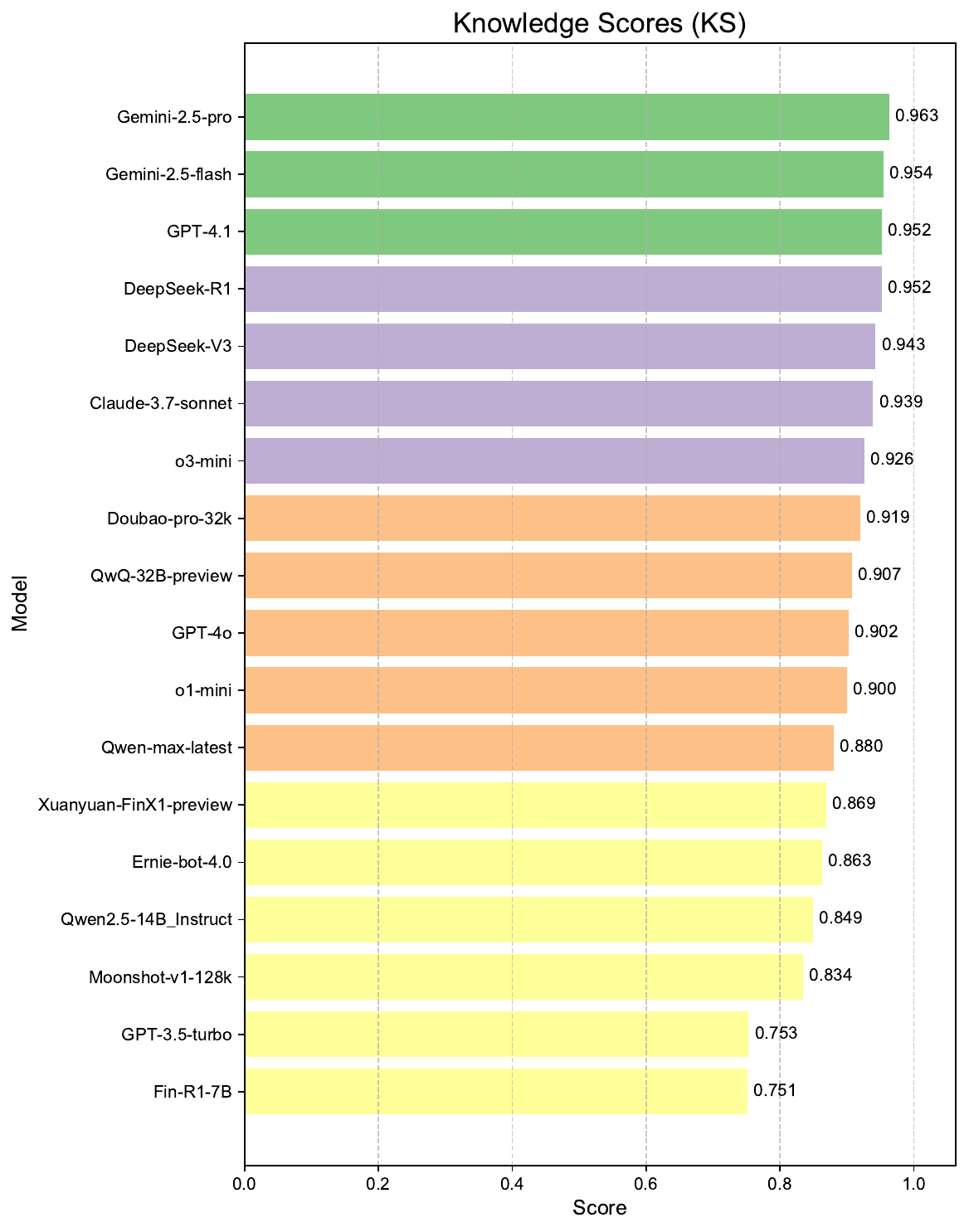}
    \caption{The knowledge score of the models.}
    \label{fig:ks}
\end{figure}

The KS measures the breadth of a LLM's knowledge coverage in the financial domain. Based on the evaluation results, models fall into four tiers, as Figure~\ref{fig:ks} shows.

Tier 1 is exclusively composed of closed-source models that exhibit exceptionally high financial knowledge coverage. Tier 2 is dominated by the top-performing open-source models. Although they rank just below Tier 1, the absolute score difference is marginal, indicating that their financial knowledge coverage is nearly on par with the leading closed-source models.

A significant performance drop-off occurs in the lower tiers. In Tiers 3 and 4, the older GPT-3.5-turbo notably outperforms other models within this bracket. At the bottom of the ranking is the specialized financial model, Fin-R1-7B, whose lower performance is primarily attributed to its significantly smaller parameter scale.

In summary, leading closed-source and top open-source reasoning models demonstrate the strongest performance in financial knowledge coverage, which is significantly influenced by model scale. While financial knowledge is a mature capability in most mainstream LLMs and no longer the primary differentiator among top models, it remains a fundamental prerequisite for high-quality reasoning.

% -----------------

\subsection{Analysis of Reasoning Score}
\begin{figure}[!t]
    \centering
    \includegraphics[width=0.9\linewidth]{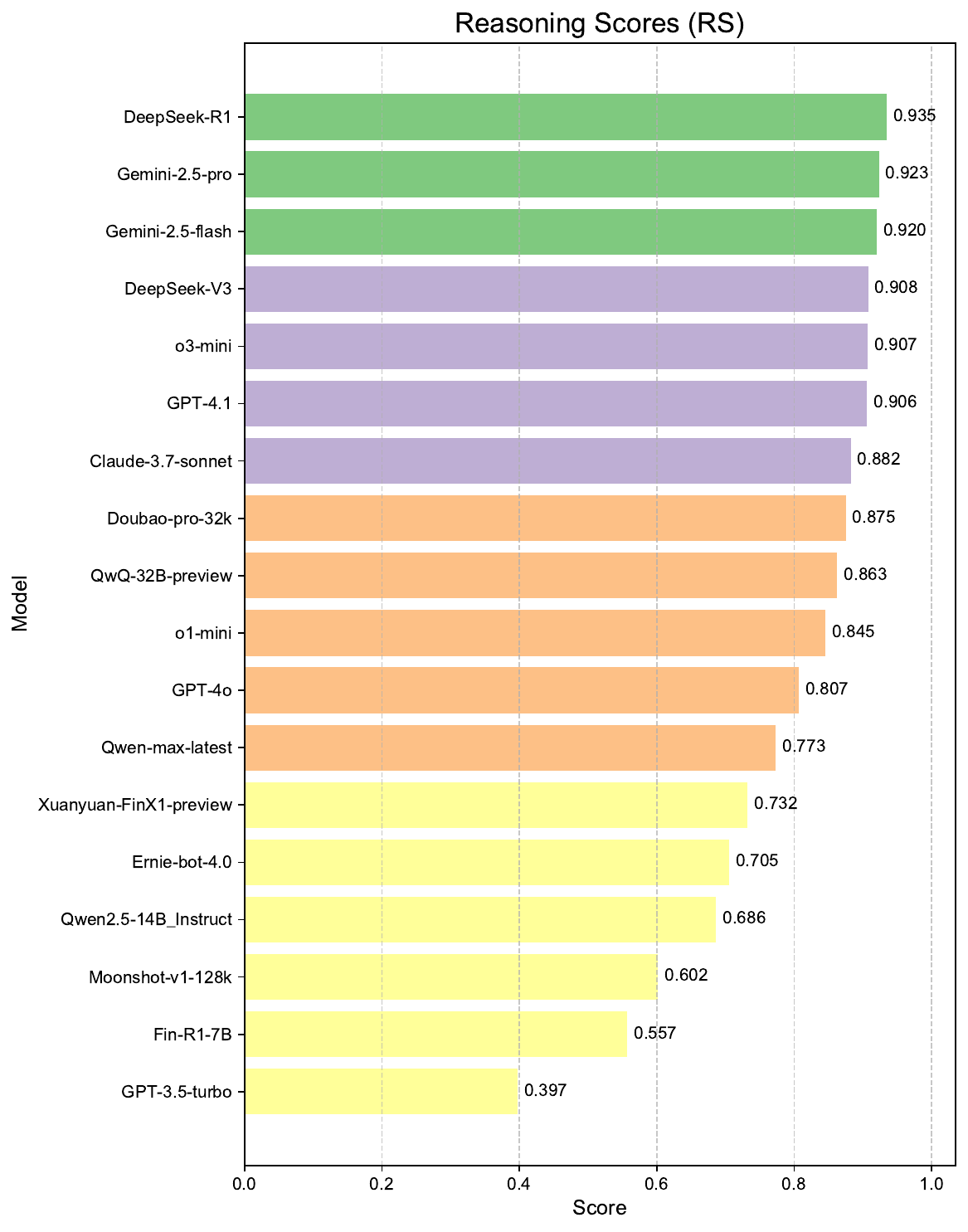}
    \caption{The reasoning score of the models.}
    \label{fig:rs}
\end{figure}

The RS is inversely related to the proportion of failures caused by incorrect reasoning steps. It reflects a model's reasoning ability. Based on this metric, the evaluated models are also divided into four tiers, as Figure~\ref{fig:rs} shows.

Tier 1 represents the pinnacle of performance, comprising reasoning-optimized models that demonstrate outstanding accuracy and logical completeness. Tier 2 includes some high-performing, non-reasoning models like GPT-4.1 and DeepSeek-V3.
A significant performance gap separates the top two tiers from the bottom two. This clear stratification underscores the need for future model development to prioritize the design and optimization of the reasoning pipeline, which is crucial for enhancing the reliability and stability of complex reasoning tasks.

% ------------------
\subsection{Analysis of Cognitive Score}

The CS provides a systematic evaluation of models' cognitive abilities based on Bloom's Taxonomy, across five dimensions, that is remembering (CS$_1$), understanding (CS$_2$), applying (CS$_3$), analyzing (CS$_4$), and evaluating (CS$_5$). As Table~\ref{tab:all_results} shows, CS scores generally exhibit a positive correlation with the KS and the RS.

\begin{figure}[!t]
    \centering
    \includegraphics[width=0.9\linewidth]{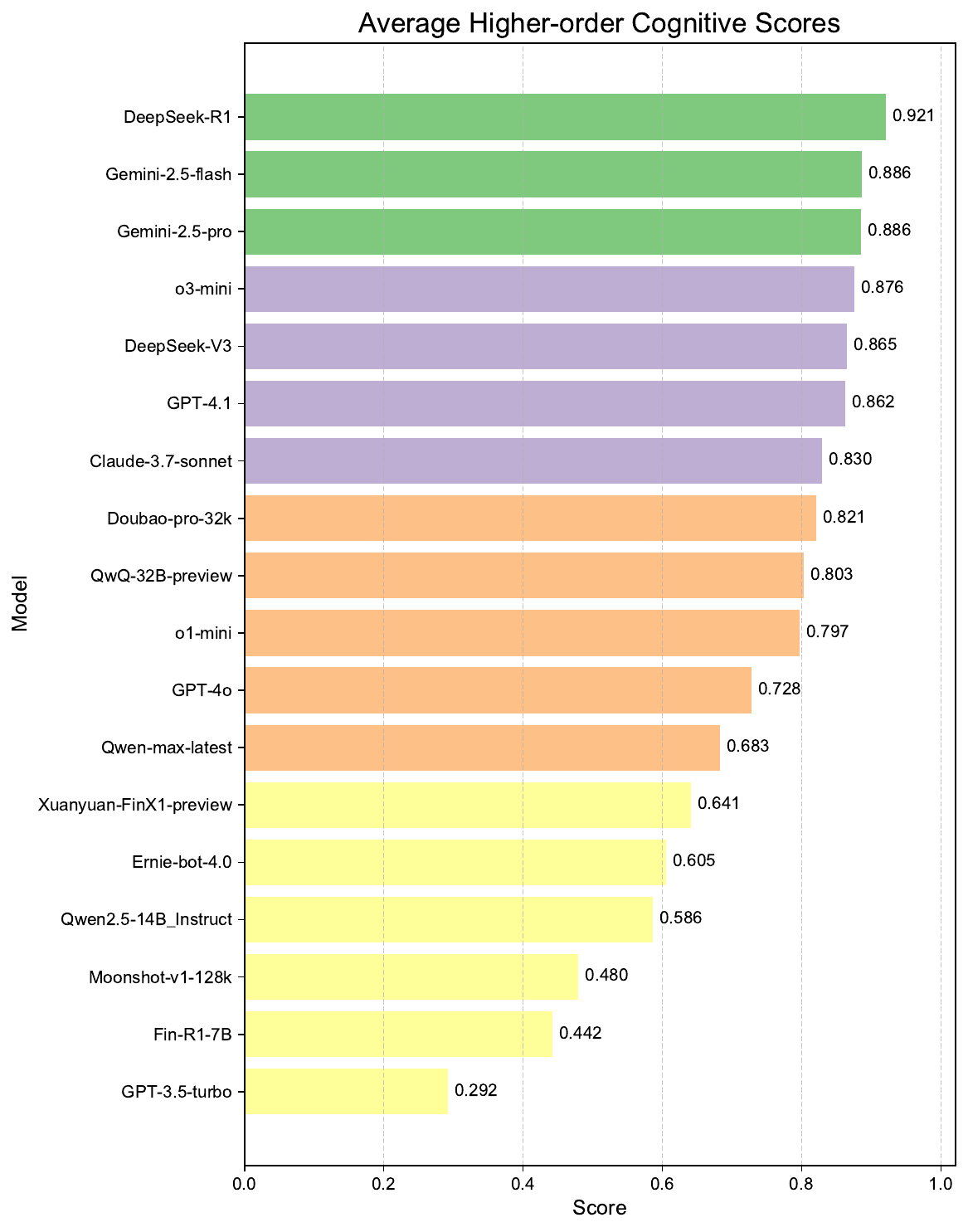}
    \caption{The average higher-order cognitive scores (CS$_3$ to CS$_5$) for the model.}
    \label{fig:cs}
\end{figure}

While most models achieve high scores at lower cognitive levels (CS$_1$: Remembering, CS$_2$: Understanding), their performance diverges significantly on higher-order tasks. These more demanding abilities—Applying (CS$_3$), Analyzing (CS$_4$), and Evaluating (CS$_5$)—reveal clear distinctions among the models. Consequently, our analysis focuses on these three dimensions. We define a primary metric, $\text{CS}{_\text{avg}}$, as the average score across these higher-order skills, and stratify the models into four performance tiers based on this metric (see Figure~\ref{fig:cs}).

Tier 1 models excel across all cognitive levels, demonstrating a distinct advantage in higher-order abilities. This tier is led by DeepSeek-R1, followed by Gemini-2.5-flash and pro.
Tier 2 models also exhibit strong higher-order cognitive skills, with performance slightly below that of Tier 1. This tier includes most general-purpose reasoning models as well as the top-performing non-reasoning models, DeepSeek-V3 and GPT-4.1.
Tiers 3 and 4 primarily consist of non-reasoning or smaller-scale models.

% -----------------------------
\subsection{Analysis of Task Accuracy}
\begin{figure}[!t]
    \centering
    \includegraphics[width=0.9\linewidth]{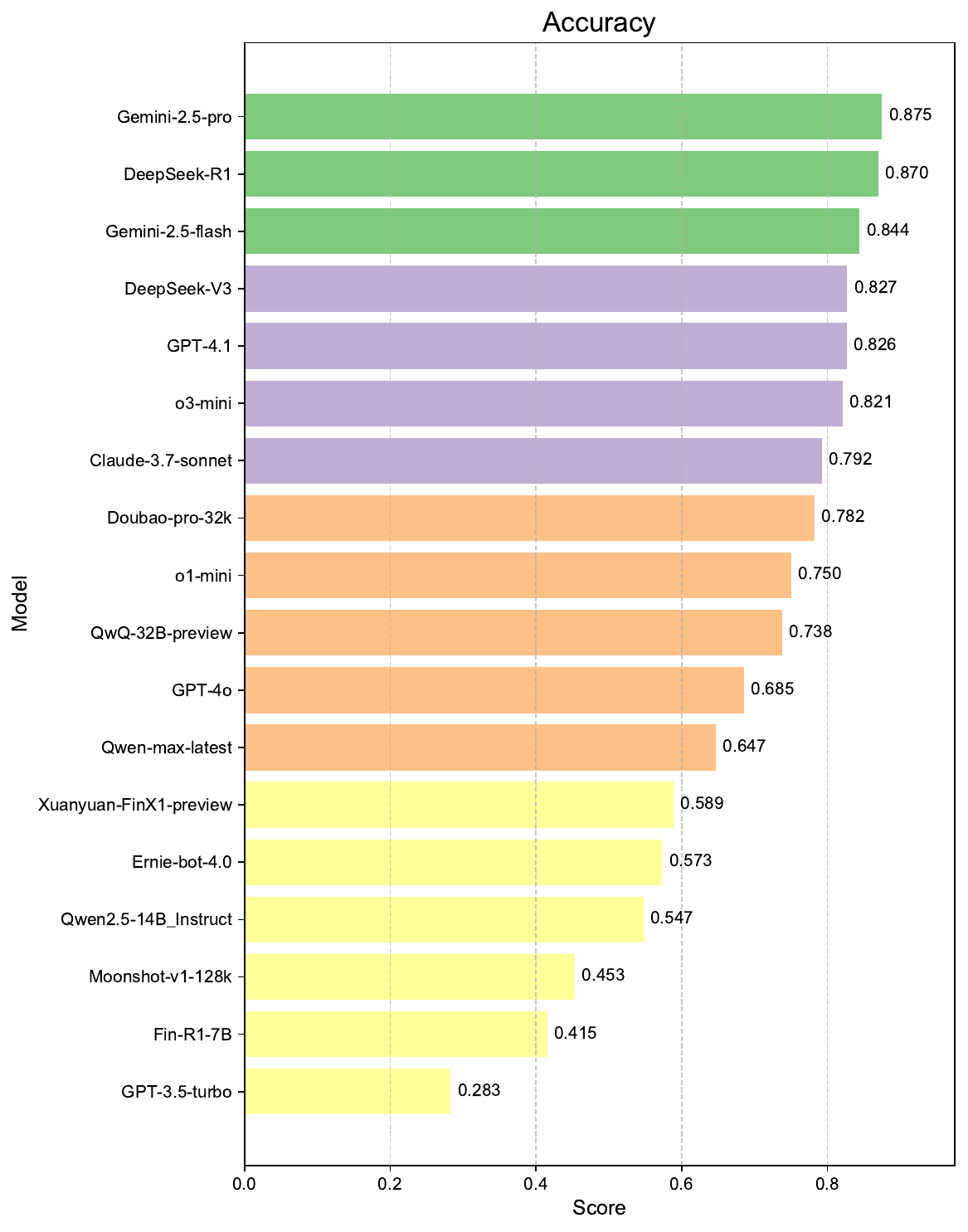}
    \caption{The accuracy of the models.}
    \label{fig:acc}
\end{figure}

Task Accuracy measures a model's direct success rate in executing reasoning tasks. Achieving high accuracy requires a synthesis of a broad knowledge base, robust reasoning capabilities, and advanced cognitive skills—particularly in application and analysis. Consequently, the performance gradient observed in Task Accuracy closely mirrors those of the RS and CS. The tiers of models based on this metric are shown in Figure~\ref{fig:acc}.

% -----------------------

\subsection{Variance Analysis}

We evaluate a model's performance stability by the standard deviation of its scores across multiple test runs. We classify stability into two categories: \textit{High Stability} (a standard deviation on the order of $10^{-4}$ to $10^{-3}$), indicating highly consistent and reproducible outputs, and \textit{Low Stability} (an order of $10^{-2}$ to $10^{-1}$), which suggests significant performance fluctuations.

Our key findings are as follows:
\begin{itemize}
    \item \textbf{Knowledge Retrieval is More Stable Than Reasoning.} For most models, the KS is consistently more stable than the RS. This is intuitive, as retrieving a stored fact is a more deterministic process for a well-trained model than performing a complex, multi-step logical deduction, which allows for greater variability.

    \item \textbf{GPT-4o is a Unique Outlier.} GPT-4o defies the general trend. Its reasoning process is remarkably stable, with an RS standard deviation of $8 \times 10^{-3}$, which is significantly more stable than its knowledge retrieval (KS standard deviation of $2 \times 10^{-2}$). We hypothesize that GPT-4o may possess a highly consistent, almost programmatic reasoning structure, while its knowledge function exhibits greater variance to adapt to diverse queries. This unusual stability profile warrants further investigation.
\end{itemize}

\section{Details of Human Evaluators and Validation Process}
\label{sec:app_labelor}
\subsection{Evaluator Qualifications and Number}
A total of 30 human experts participated in our validation effort. All experts are postgraduate students with academic backgrounds in finance, economics, or statistics, ensuring they possess an accurate understanding of the relevant professional terminology, fundamental concepts, and practical scenarios.

The entire validation process was conducted on a professional annotation platform provided by a leading technology company to ensure procedural standardization and data security.

\subsection{Quality Control Mechanism}
To guarantee the reliability of our validation results, we implemented a multi-stage quality control process. First, we randomly sampled 10\% of the dataset. Each sample was independently validated by 2 experts to ensure consistency. Following this cross-validation, we organized a team of 3 senior experts to conduct a final quality check on a random 10\% of the already-validated sample (amounting to a final check on 1\% of the total dataset). This final step was designed to ensure the quality and uniformity of the standards applied during the cross-validation stage.

\subsection{The Validation Process}
The experts' validation work was divided into three strict, sequential stages:
\paragraph{Stage 1: Question and Knowledge Point Validation} 
In this initial stage, experts were only shown the question and its associated knowledge points. They were tasked with the following checks:
\begin{itemize}
    \item \textbf{Question Validity}: Is the question relevant to a realistic financial scenario? Is professional terminology used correctly? Are the numerical values within a reasonable range? Does the question have a single, definitive answer?

    \item \textbf{Knowledge Point Relevance}: Are the tagged knowledge points accurate and comprehensive? Is the naming of the knowledge points consistent with standard terminology in mainstream textbooks?
\end{itemize}

\paragraph{Independent Answering}
After confirming the quality of the question, experts were required to solve the problem independently, without reference to any provided solution. The goal of this step was to obtain a high-quality, unbiased human answer to serve as a benchmark for subsequent comparisons.

\paragraph{Stage 3: Reasoning Steps and Cognitive Labels Validation}
Finally, the system presented the experts with the answer, the step-by-step solution, and the Bloom's Taxonomy cognitive label from our dataset. The experts were required to perform the following checks:
\begin{itemize}
    \item \textbf{Answer and Solution Process Verification}: First, they compared their own answer to the one in the dataset. If the answers did not match, the sample was immediately flagged as ``unqualified''. If the answers matched, they proceeded to meticulously review the solution steps provided in the dataset, assessing whether the logic was clear, the steps were reasonable, and the calculations were correct.

    \item Cognitive Label Accuracy Check: Based on the predefined verb list corresponding to each cognitive level (as defined in Figure~\ref{fig:annotate_cognitive_label}), the experts had to judge whether the cognitive label assigned to the question was accurate.
\end{itemize}
% -----------------------
\section{License and Usage Constraints}
The released dataset is distributed under the Creative Commons Attribution-NonCommercial 4.0 International License (CC BY-NC 4.0).

The dataset is only for for evaluating LLMs in non-commercial academic research. The dataset is explicitly not authorized for use in training or fine-tuning machine learning models , including pre-training, instruction-tuning, or reinforcement learning stages. Access conditions ensure that all derived data products remain confined to research contexts, with no transfer or application permitted in industrial, governmental, or other operational domains.

\section{AI Assistants Usage Disclosure}
This study did not employ any AI assistants during the research design, data analysis, or coding phases. During manuscript preparation, the authors exclusively utilized Google Gemini for grammatical refinement and stylistic polishing. No AI-generated content was incorporated into the methodology and results, ensuring the work's originality and human-driven intellectual integrity.

\end{document}